\documentclass[10pt,journal,compsoc]{IEEEtran}
\usepackage{times}
\usepackage{epsfig}
\usepackage{graphicx}
\usepackage{amsmath}
\usepackage{amssymb}
\usepackage{helvet}
\usepackage{courier}
\usepackage{upgreek}
\usepackage{color}
\usepackage{url}
\usepackage{booktabs}
\usepackage{multirow}
\usepackage{diagbox}
\usepackage{bm}
\usepackage{cases}
\usepackage[colorlinks,citecolor=black,linkcolor=black]{hyperref}
\usepackage{array}
\newcommand{\PreserveBackslash}[1]{\let\temp=\\#1\let\\=\temp}
\newcolumntype{C}[1]{>{\PreserveBackslash\centering}p{#1}}

\ifCLASSOPTIONcompsoc
  \usepackage[nocompress]{cite}
\else
  \usepackage{cite}
\fi

\ifCLASSINFOpdf
\else
\fi

\begin{document}
\title{AGO-Net: Association-Guided 3D Point Cloud \\ Object Detection Network}

\author
{Liang~Du,
 Xiaoqing~Ye,
 Xiao~Tan,
 Edward~Johns,
 Bo~Chen,\\
 Errui~Ding,
 Xiangyang~Xue,
 Jianfeng~Feng
 \IEEEcompsocitemizethanks{
 \IEEEcompsocthanksitem L. Du and J. Feng are with the Institute of Science and Technology for Brain-Inspired Intelligence, Fudan University, Shanghai, China, Key Laboratory of Computational Neuroscience and Brain-Inspired Intelligence (Fudan University), Ministry of Education, China, MOE Frontiers Center for Brain Science, Fudan University, Shanghai, China, Zhangjiang Fudan International Innovation Center.
 J. Feng is also with the Fudan ISTBI-ZJNU Algorithm Centre for Brain-inspired Intelligence, Zhejiang Normal University, Jinhua, China.
 \IEEEcompsocthanksitem X. Ye, X. Tan, and E. Ding are with Baidu Inc., China.
 \IEEEcompsocthanksitem E. Johns is with the Robot Learning Lab, Imperial College London, UK.
 \IEEEcompsocthanksitem B. Chen is with FAW (Nanjing) Technology Development Co., Ltd, China.
 \IEEEcompsocthanksitem X. Xue is with the School of Computer Science, Fudan University, China.
 \IEEEcompsocthanksitem Email: duliang@mail.ustc.edu.cn, jffeng@fudan.edu.cn}
}
\IEEEtitleabstractindextext{

 \begin{abstract}
 The human brain can effortlessly recognize and localize objects, whereas current 3D object detection methods based on LiDAR point clouds still report inferior performance for detecting occluded and distant objects: the point cloud appearance varies greatly due to occlusion, and has inherent variance in point densities along the distance to sensors. Therefore, designing feature representations robust to such point clouds is critical. Inspired by human associative recognition, we propose a novel 3D detection framework that associates intact features for objects via domain adaptation. We bridge the gap between the perceptual domain, where features are derived from real scenes with sub-optimal representations, and the conceptual domain, where features are extracted from augmented scenes that consist of non-occlusion objects with rich detailed information. A feasible method is investigated to construct conceptual scenes without external datasets. We further introduce an attention-based re-weighting module that adaptively strengthens the feature adaptation of more informative regions. The network's feature enhancement ability is exploited without introducing extra cost during inference, which is plug-and-play in various 3D detection frameworks. We achieve new state-of-the-art performance on the KITTI 3D detection benchmark in both accuracy and speed. Experiments on nuScenes and Waymo datasets also validate the versatility of our method.
\end{abstract}

 \begin{IEEEkeywords}
  3D object detection, domain adaptation, associative recognition, point cloud, neural network, autonomous driving
 \end{IEEEkeywords}}

\maketitle
\IEEEdisplaynontitleabstractindextext
\IEEEpeerreviewmaketitle

\section{Introduction}
\label{sec:intro}
\IEEEPARstart{O}{bject} detection \cite{fidler20123d, luo2018fast, liang2018deep, ye2021} is drawing increasing attention from both academia and industry due to its critical role in robotics and autonomous driving \cite{geiger2012we}. Despite the recent success of image-based 2D object detection \cite{ren2015faster, centernet}, 3D object detection from LiDAR-based point clouds remains an under-developed research area, and a highly challenging problem due to object occlusion and variance in point distributions. The density of a point cloud decreases as the distance to a LiDAR camera increases, which causes an apparent density discrepancy. Moreover, some parts of objects are invisible due to occlusion or low point cloud occupancy. In this situation, 3D detection is unreliable and error-prone because the network fails to extract sufficient features.

\begin{figure}[t]
 \begin{center}
  \includegraphics[width=7.5cm]{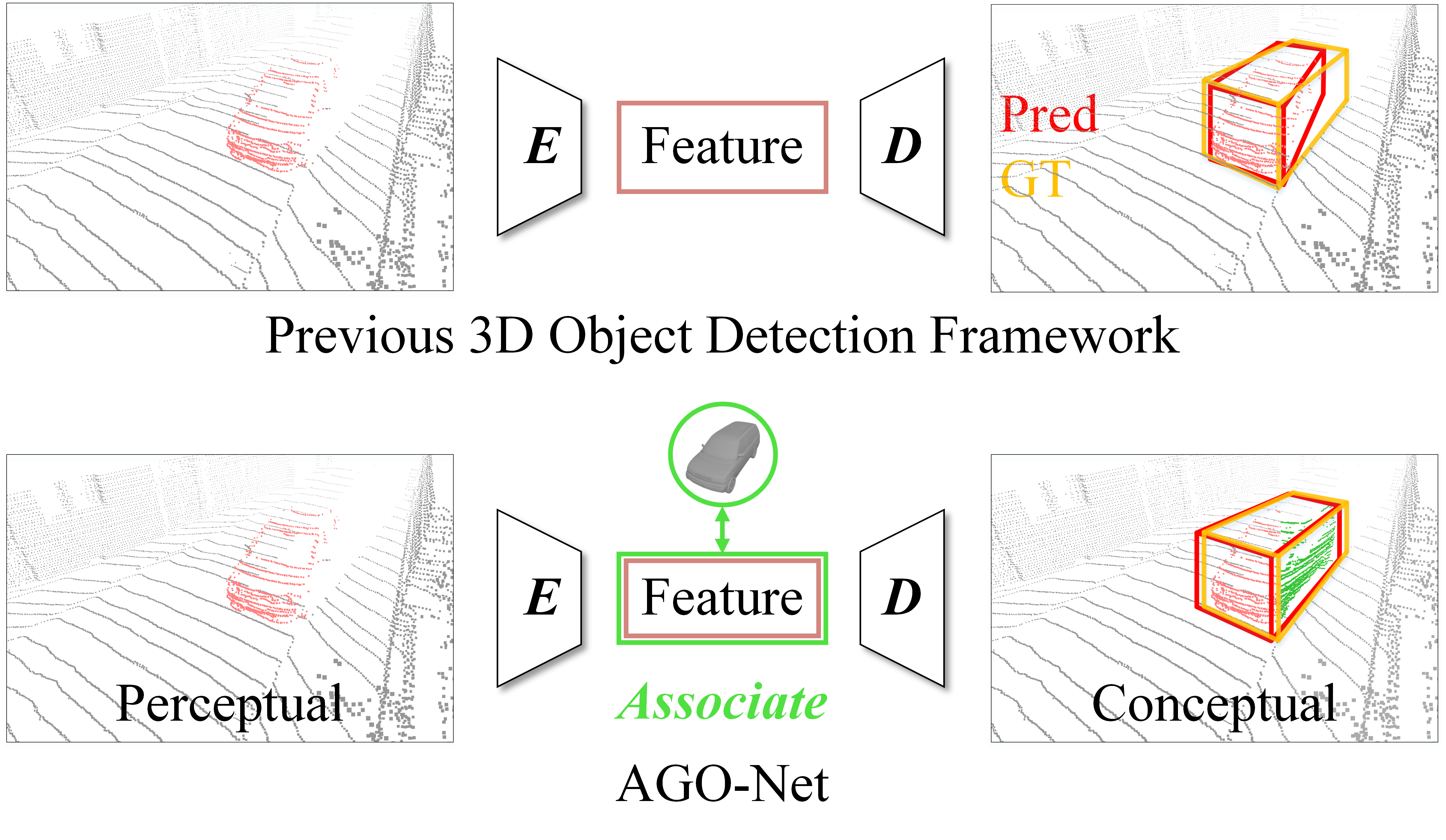}
 \end{center}
 \caption{
  Illustration of the standard 3D object detection framework and the proposed associative-recognition-inspired AGO-Net. The encoder and the decoder are denoted by $E$ and $D$, respectively. Compared with previous approaches that solely utilized the encoded features of real-world point clouds for 3D object detection, our AGO-Net builds up the association between incomplete perceptual features of real-world objects and more complete features of corresponding class-wise conceptual models via domain adaptation. AGO-Net mimics the conceptual association of the human brain when perceiving objects, and fundamentally exploits the network capability of the feature enhancement.
 }
 \label{fig1}
\end{figure}

Recently, many cutting-edge 3D object detection approaches \cite{yang2019std, wang2019range} have attempted to solve this problem. In STD \cite{yang2019std}, PointsPool is introduced to convert sparse intermediate point features to a more compact voxel representation. Nevertheless, for far-range objects, the voxelization operation does not facilitate the network in generating sufficient and robust features because most voxels are empty. A generative adversarial network (GAN) is leveraged by \cite{wang2019range} to force the network to generate consistent features between far-range and near-range objects. However, this method aims at range domain adaptation and ignores object-wise constraints, such as shapes and the viewing perspective of the objects. PV-RCNN \cite{Shi_2020_CVPR} adopts both voxel-based features and point-based features of point clouds and obtains more discriminative representations at the cost of feature extraction.

Previous works like \cite{hassabis2017neuroscience, du20203dcfs} have proven that exploiting the biological and psychological plausibility of the human brain to build up intelligent systems brings significant performance gains to various tasks. For the human psychological model, perceiving 3D objects is carried out in the form of a conceptual association process, which includes two vital stages \cite{carlesimo1998associative}: (I) a ``viewer-centered'' feature representation stage, where the features of perceptual objects are presented from the viewer's perspective, and the visible features might lack structure and detail information due to occlusion and large distance, and (II) an ``object-centered'' stage, where the object's features are enhanced by the conceptual model of the same category retained in the memory. This association process helps map from an object to a view-invariant 3D conceptual model, which is termed as associative recognition.

Considering the previously mentioned issues, we propose a brain-inspired 3D object detection framework that associates weak perceptual features of occluded and distant objects with complete conceptual models' features that are more favorable for detection. As depicted in Figure \ref{fig1}, we select structurally-complete 3D models as conceptual guidance and adapt object-wise features from the perceptual domain to the conceptual domain during training following a transfer learning paradigm. Comparing with the conference version~\cite{Du_2020_CVPR}, there are four main differences.
(1) For network architecture, instead of directly performing domain adaptation on the encoded feature, we decouple the features and only adapt the regression feature according to the dynamic attention map generated by the classification feature (SC-reweight), which achieves superior performance.
(2) For the conceptual domain construction, an object completion method is proposed to maintain the original structure of each instance. 
(3) For the performance and experiments, our AGO-Net ranked $1^{st}$ among all existing one-stage detectors in the KITTI test benchmark at the time of submission. In addition to the KITTI experiments mainly on the ``Car'' category in ~\cite{Du_2020_CVPR}, we further perform multi-class experiments on more challenging datasets such as nuScenes and Waymo.
(4) For portability, we have also conducted experiments integrating our method into other state-of-the-art (SoTA) 3D detection frameworks, which validates its generalizability.

The main contributions are summarized as follows:
\begin{itemize}
 \item We propose a 3D detection framework that associates real-world perceptual features with more complete features from conceptual models, which strengthens the feature robustness, especially for occluded and distant objects.
 \item We investigate a novel attention-based adaptation module that adapts more critical features from the real-world perceptual domain to the constructed conceptual domain.
 \item We present a feasible approach to generate the proposed conceptual scenes without requiring external datasets.
 \item Our method achieves new SoTA 3D object detection performance on the KITTI dataset in both accuracy and speed (25 FPS). More evaluations on the nuScenes and Waymo datasets further validate the effectiveness.
\end{itemize}
\section{Related work}
\label{sec:related}

\begin{figure*}[t]
 \begin{center}
  \includegraphics[width=15.8 cm]{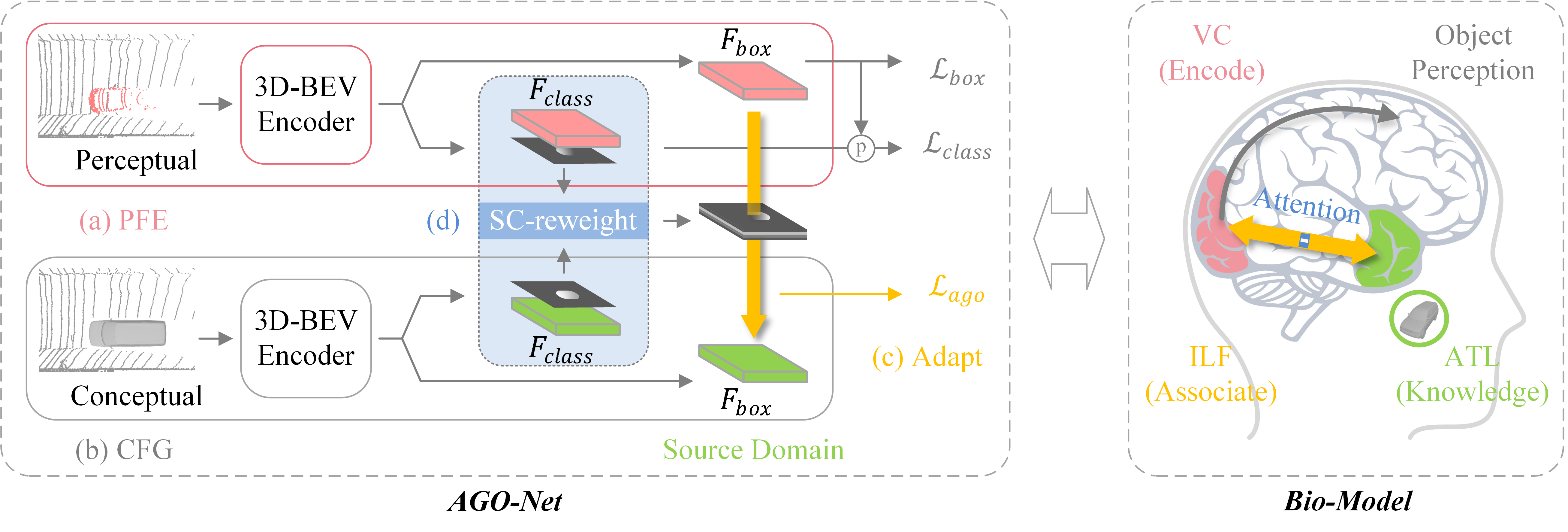}
 \end{center}
 \caption{ Overview of the AGO-Net framework and its underlying biological model. AGO-Net consists of four parts:
  (a) The perceptual feature encoder (PFE), which encodes perceived information in a manner similar to that employed by the visual cortex (VC) in the human brain for object perception, as shown in the right figure. Various 3D detection methods can be utilized as alternatives to the 3D-BEV encoder, which encodes a 3D point cloud scene to a compact BEV representation.
  (b) The conceptual feature generator (CFG) generates the features based on the constructed conceptual scene, which provides conceptual association guidance for the PFE. This detachable auxiliary network is directly discarded during the inference time. (c) The feature adaptation from the real-world perceptual domain to the constructed conceptual domain mimics the knowledge association and retrieval process in the human brain. The large yellow arrow shows the adaptation direction. (d) The \textbf{S}patial and \textbf{C}hannel-wise attention-aware reweighting module (SC-reweight) aids the siamese network in adapting more crucial features based on the difference feature map of the conceptual and perceptual classification features. The loss terms of AGO-Net include the localization (XYZ), scale (WHL), classification, and rotation losses as well as our proposed association loss $\mathcal{L}_{ago}$. A detailed explanation of the Bio-model is provided in Sec.\ref{2}. }
 \label{fig2}
\end{figure*}

\noindent \textbf{3D point cloud object detection.}
On the one hand, recent 3D object detection approaches usually include four types: multi-view-based, voxel-based, point-based, and graph-based methods \cite{Shig_2020_CVPR}. On the other hand, if divided by whether proposals and post-refinement are required, 3D detection approaches can be divided into one-stage and two-stage methods.

The multi-view approaches such as \cite{chen2017multi, ku2018joint, yang2018pixor, yang2018hdnet} project LiDAR point clouds to the image plane \cite{li20173d} or the bird’s eye view (BEV) to encode features. Zhou et al. \cite{zhou2019endtoend} proposed a multi-view fusion framework that utilizes the complementary information from BEV and perspective view.

The voxel-based methods utilize voxel representations for point cloud encoding. Vote3Deep \cite{engelcke2017vote3deep} and 3DFCN \cite{li20173d} evenly discretize point clouds on grids and apply 3D convolutional networks to address point clouds for object detection. Zhou et al. proposed VoxelNet to learn discriminative features from sparse point clouds \cite{zhou2018voxelnet}. The following works \cite{Graham2017Submanifold, yan2018second} improve VoxelNet by adopting 3D sparse convolutional layers to save computational resources. Lang et al. introduced PointPillars \cite{lang2019pointpillars}, which utilizes PointNets \cite{qi2017pointnet} to learn a representation of point clouds that are organized in vertical columns (pillars), which boosts the inference speed. Vora et al. introduced PointPainting \cite{Vora_2020_CVPR}, which projects predicted 2D pixel-wise semantic labels to point clouds as additional prior knowledge for the network. SA-SSD \cite{He_2020_CVPR} employs an auxiliary network that converts the convolutional features back to point-level representations. The point-level supervision guides the convolutional features to be aware of the structure. Cheng et al. proposed PPBA \cite{cheng2020improving}, which learns to optimize data augmentation strategies by narrowing the search space and adopting the best parameters discovered in previous iterations.

The point-based method F-PointNet \cite{qi2018frustum} is the first to exploit raw point clouds for 3D object detection. Frustum proposals generated by off-the-shelf 2D object detectors are utilized as candidate boxes, and final detection outputs are regressed based on interior points \cite{xu2018pointfusion}. Shi et al. presented PointRCNN \cite{shi2019pointrcnn}, which directly encodes the raw point cloud to predict 3D proposal regions in a bottom-up manner based on the segmentation labels from 3D bounding box annotations. 3DSSD \cite{Yang_2020_CVPR} is the first one-stage point-based method that removes the refinement module in all existing point-based methods and introduces a fusion sampling strategy for retaining sufficient interior points of different objects for detection.

In summary, the voxel-based methods such as \cite{yan2018second, wang2019range, He_2020_CVPR}, that parse the 3D scene into a compact representation and adopt straight-through CNN networks, rather than a coarse-to-refine method with proposals and refinement, can be regarded as one-stage methods. One-stage approaches gain efficiency while losing some precision due to the progressively downscaled feature maps. The point-based methods like \cite{shi2019pointrcnn, yang2019std, Shi_2020_CVPR} exploit spatial features of the raw points based on the region proposal generated by the first stage, but they sacrifice inference speed for accuracy.

\noindent \textbf{Deep neural network-based transfer learning.}
Transfer learning \cite{pan2009survey} is a machine learning paradigm that captures the notion of knowledge transfer between related task domains in computer vision \cite{hoffman2014lsda, Du_2019_ICCV} and natural language processing (NLP) \cite{collobert2011natural}. Traditional domain adaptation in transfer learning alleviates the influence of the distribution mismatch between the source domain and the target domain (e.g., training and testing data, synthetic and real-world data, respectively) so that the generalizability of the network can be boosted for the target domain data. Maximum Mean Discrepancy (MMD) introduced by \cite{gretton2006a} is a common domain distance metric. In addition to handcrafted distribution distance metrics, deep neural networks are leveraged to perform transfer learning. The learning-based methods can extract high-level representations that disentangle different explanatory factors of variations in the data \cite{bengio2013representation}. Invariant factors that underlie different populations are also manifested, which transfer well from source domain tasks to target domain tasks \cite{yosinski2014transferable}. \cite{Liu2019SKD} utilized feature distillation to bridge the domain gap between the simulated scene and real-world scene for semantic segmentation. For image classification, \cite{attentionx2} presented Attention Transfer (AT) that squeezes intermediate feature maps into a single channel via an attention mask for feature distillation. 

\section{Methodology}
\label{sec:method}
In this section, we introduce our association-guided 3D object detection network. As a result of the inevitable occlusion or large distance, many foreground objects contain only a few points, which makes it difficult for standard networks to perform classification and localization. Therefore, the perceptual domain corresponds to the real-world scene results that are not satisfactory due to the variance in the point cloud appearance. Alternatively, we can create a conceptual domain that learns from 3D conceptual models with more complete structures to guide feature learning in challenging cases. The conceptual features are discriminative enough for object classification and localization since they are extracted from point clouds with the finest details. If the domain gap between the perceptual and conceptual features is diminished, the network will have the capability to associate stronger features for objects adaptively.

\begin{figure}[htb]
 \begin{center}
  \includegraphics[width=8.2 cm]{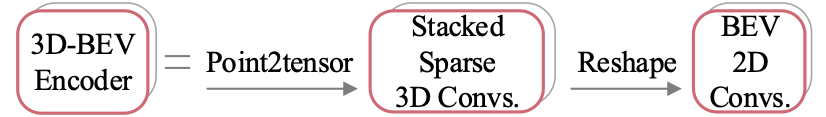}
 \end{center}
 \caption{ One instantiated network of the PFE (red) / CFG (grey). }
 \label{fig3}
\end{figure}

Figure \ref{fig2} depicts the overall framework of AGO-Net, which includes four parts: (a) the perceptual feature encoder (PFE) to encode real-world target domain features for 3D object detection; (b) the conceptual feature generator (CFG) to generate the source domain features from corresponding scenes reconstructed by the conceptual models; (c) the perceptual-to-conceptual feature domain adaptation; and (d) the \textbf{S}patial and \textbf{C}hannel-wise loss reweighting module (SC-reweight) based on the classification features from both perceptual and conceptual scenes, which forces the network to strengthen the adaptation for regions that have much more critical information. Our approach to constructing conceptual scenes without involving external data is described in Sec.\ref{4}. The network architecture and its underlying biological model that supports our motivation are also detailed in Sec.\ref{2}. In Sec.\ref{3}, we further explain the training process of AGO-Net.

\subsection{Network of AGO-Net}
\label{1}
Our AGO-Net can employ various 3D point cloud detection approaches, such as \cite{yan2018second, lang2019pointpillars, He_2020_CVPR}, etc., as its sub-networks PFE and CFG. We adopt a one-stage architecture based on sparse convolutional layers to instantiate one PFE and one CFG.

\noindent \textbf{The PFE to encode perceptual features.}
The PFE serves as the standard feature encoder of the real scenes, which is shown in Figure \ref{fig2} and colored in red. Following \cite{He_2020_CVPR}, our AGO-Net first obtains the input point features by discretizing the point coordinates to input tensor indices and then adopts 3D sparse convolution \cite{Graham2017Submanifold} for feature extraction. Specifically, for the point cloud quantization, we denote the coordinates of the $i_{th}$ point in each scene as $(C^i_x, C^i_y, C^i_z)$. To discretize the coordinates into integer indices for the input tensor, we define $(S_x, S_y, S_z)$ as the quantization step along the X-, Y-, and Z-axes. Then, the index of each point is given by:
\begin{equation}
 \begin{array}{r}
  P(I^i_x, I^i_y, I^i_z) = P(\lfloor\frac{C^i_x}{S_x}\rfloor, \lfloor\frac{C^i_y}{S_y}\rfloor, \lfloor\frac{C^i_z}{S_z}\rfloor),
 \end{array}
 \label{eq0}
\end{equation}
where $\lfloor \cdot \rfloor$ denotes the floor operation. Based on the obtained index, each point can be assigned to an entry of the network input tensor. If multiple points share the same index, then the entry is simply overwritten with the latest point.

As the red part depicted in Figure \ref{fig2}, the network architecture of the PFE can be formulated as follows:
\begin{equation}
 \left\{
 \begin{array}{lr}
  F_{class}^{\mathcal{P}}, F_{box}^{\mathcal{P}} = \xi^\mathcal{P}(I^{\mathcal{P}}) \\
  O_{class}^{\mathcal{P}}, O_{box}^{\mathcal{P}} = \mathcal{H}_{class}^{\mathcal{P}}(F_{class}^{\mathcal{P}}), \mathcal{H}_{box}^{\mathcal{P}}(F_{box}^{\mathcal{P}}),
 \end{array}
 \label{eq_PFE}
 \right.
\end{equation}
where $\xi^\mathcal{P}$ and $I^{\mathcal{P}}$ denote the 3D-BEV encoder and the perceptual input tensor, respectively. $F_{class}^{\mathcal{P}}$ and $F_{box}^{\mathcal{P}}$ are the output features of the 3D-BEV encoder for object classification and 3D bounding box regression. $\mathcal{H}_{class}^{\mathcal{P}}$ and $\mathcal{H}_{box}^{\mathcal{P}}$ represent sibling convolutions to decode the features for the output predictions $O_{class}^{\mathcal{P}}$ and $O_{box}^{\mathcal{P}}$, respectively. Note that $\xi$ can be instantiated by a variety of 3D object detectors.

\begin{figure}[t]
 \begin{center}
  \includegraphics[width=8.6cm]{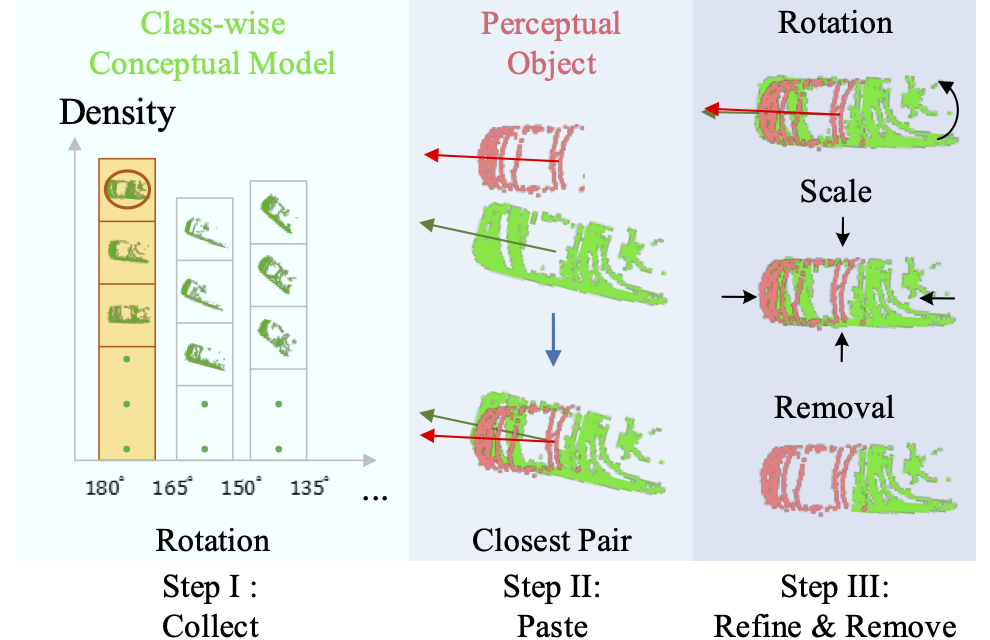}
 \end{center}
 \caption{ The generation process of the conceptual model and scene. }
 \label{fig4}
\end{figure}

As shown in Figure \ref{fig3}, the ``Sparse 3D Convs.'' in the first red box encodes the point cloud feature into a high-dimensional feature. The feature is then reshaped and converted to a BEV representation via concatenating the feature along the height dimension of the scene into one channel. After that, ``BEV 2D Convs.'' in the third box encodes the compacted features into $F_{class}^{\mathcal{P}}$ and $F_{box}^{\mathcal{P}}$. Finally, the network uses two heads $H_{box}$ and $H_{class}$ to decode two features for box regression and classification, respectively.

To alleviate the common issue of misalignment between the classification confidence maps and the corresponding predicted bounding boxes for 3D object detection, we apply the part-sensitive warping (PSWarp) method following \cite{He_2020_CVPR} to align the classification confidences with the bounding boxes via a spatial transformation of the feature map, as shown in Figure \ref{fig2} (the grey circled ``p'' at the end of the network). In terms of efficiency, we choose PSWarp instead of the alternative method PSRoIAlign \cite{dai2016r-fcn:}, since \cite{He_2020_CVPR} has proven that the performance of PSWarp is comparable to the performance of PSRoIAlign, while PSWarp takes only 1/10 the runtime of PSRoIAlign.

\noindent \textbf{The CFG to generate conceptual features.}
With the same architecture of the PFE, the siamese network CFG generates the conceptual feature that served as the source domain feature. As the grey part depicted in Figure \ref{fig2}, the CFG is formulated as follows:
\begin{equation}
 \left\{
 \begin{array}{lr}
  F_{class}^{\mathcal{C}}, F_{box}^{\mathcal{C}} = \xi^\mathcal{C}(I^{\mathcal{C}}) \\
  O_{class}^{\mathcal{C}}, O_{box}^{\mathcal{C}} = \mathcal{H}_{class}^{\mathcal{C}}(F_{class}^{\mathcal{C}}), \mathcal{H}_{box}^{\mathcal{C}}(F_{box}^{\mathcal{C}}).
 \end{array}
 \label{eq_CFG}
 \right.
\end{equation}
Eq. \ref{eq_CFG} is similar to Eq. \ref{eq_PFE}, except that subscript $\mathcal{C}$ in Eq. \ref{eq_CFG} denotes the conceptual domain, whereas subscript $\mathcal{P}$ in Eq. \ref{eq_PFE} represents the perceptual domain. We first train the CFG alone. The conceptual dataset is derived from the real-world dataset and is integrated with the conceptual models. Specifically, conceptual scenes denote object cloud instances that have complete appearance to some degree. Original point clouds with enough points and an intact structure are simply included. The conceptual model is a more complete point cloud for each object, such as a CAD model from external resources, or a surrogate model with more informative knowledge originating from the same dataset, i.e., self-contained. The format of the conceptual model is not specifically restrained.

After the CFG is well trained, the parameters of the CFG are frozen to provide stable feature guidance for further domain adaptation, and the decoding part of the CFG is removed, as shown in Figure \ref{fig2}. During the training of the entire AGO-Net, the perceptual scene and its corresponding conceptual scene are simultaneously fed into the PFE and CFG, respectively. The PFE is encouraged to generate more complete features based on the perceptual feature of the sparse and partial-visible point clouds under the guidance of the CFG. During the inference time, we simply remove the CFG from our framework.

\noindent \textbf{Self-contained method to build conceptual scenes.}
\label{4}
The conceptual models that are utilized to construct the conceptual scenes can be fabricated via various methods, such as 3D CAD models and render-based techniques. In this paper, instead of appealing to costly external resources, we propose a self-constrained scene-constructing method to adopt surrogate models with more structural information that originated from the same dataset. Since 3D point cloud objects of the same category usually have a similar shape and scale, we collect the free-of-charge ground-truth objects with more complete structures as the candidate conceptual models.

As shown in Figure \ref{fig4}, the generation process of the conceptual scene contains three steps: (1) Objects are divided into $M$ groups according to their rotation angles. In each group, we choose the top $K\%$ objects with the most points as our conceptual models. (2) For each object that is not chosen as the conceptual model, we choose the conceptual model with the minimum average closest point distance from the object as its correspondence. (3) The scale and rotation are further refined, and the added points within the small neighborhood of original points (distance $<= \vartheta$) are removed to maintain the original structural information of the perceptual object. The conceptual scene is composed by pasting the conceptual models into the incomplete original point cloud.

\noindent \textbf{Domain adaptation-based feature association.}
\label{domain}
During the training process, the perceptual feature encoded by the PFE is learned to map to the conceptual feature generated by the CFG via domain adaptation, which establishes feature association. AGO-Net optimizes an additional loss term of domain distance to guide the network to generate domain-invariant source features based on the target features. In other words, we apply a much better feature representation extracted from the conceptual scene with a more complete point cloud structure as high-dimensional supervision of the 3D detection network.

\begin{figure}[htb]
 \begin{center}
  \includegraphics[width=8.8cm]{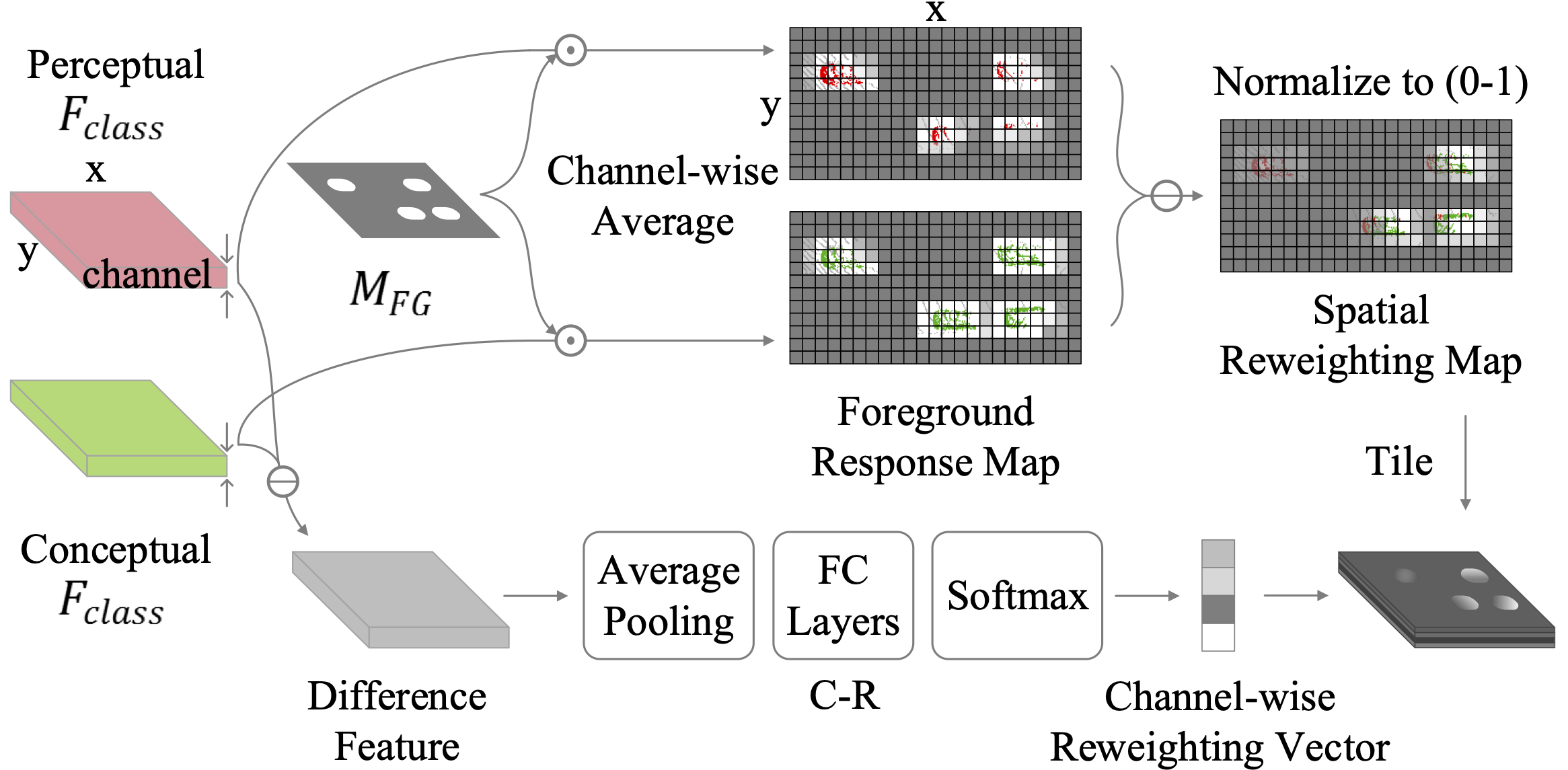}
 \end{center}
 \caption{
  Illustration of the attention-based reweighting module SC-reweight. We average the classification features along the channel of both perceptual scenes and conceptual scenes to obtain the response maps. The difference map of the response maps is calculated and multiplied by the foreground mask to obtain the spatial reweighting map. The regions with added point clouds, which contain more critical structural information and have a higher difference response, are given more attention during domain adaptation. To obtain the channel-wise reweighting vector, we utilize the difference feature map between perceptual features and conceptual features for classification as the input of a sub-network ``C-R'' in the SC-reweight. The channel-wise reweighting vector is calculated by the C-R and multiplied in a channel-wise manner with the tiled spatial reweighting map for adaptation.
 }
 \label{fig5}
\end{figure}

Since the locations of both the real-world scanned objects and the conceptual models are spatially aligned, the smooth-$L1$ distance between these two features is directly optimized for domain adaptation. The parameters of the PFE are tuned to force perceptual objects to generate features that resemble more informative conceptual features. We further constrain the domain adaptation regions to encourage the network to focus on the feature association of the foreground objects, which means that the adaptation is restricted to foreground pixels. As illustrated in Figure \ref{fig5}, the foreground object mask $\mathcal{M}_{FG}$ is obtained by reprojecting the 3D object ground truth bounding boxes to the bird’s-eye view (BEV) map, which will be further explained in our supplemental material. We then downsample the mask map to match the scale of the perceptual and conceptual features.

\begin{figure*}[htb]
 \begin{center}
  \includegraphics[width=16.3cm]{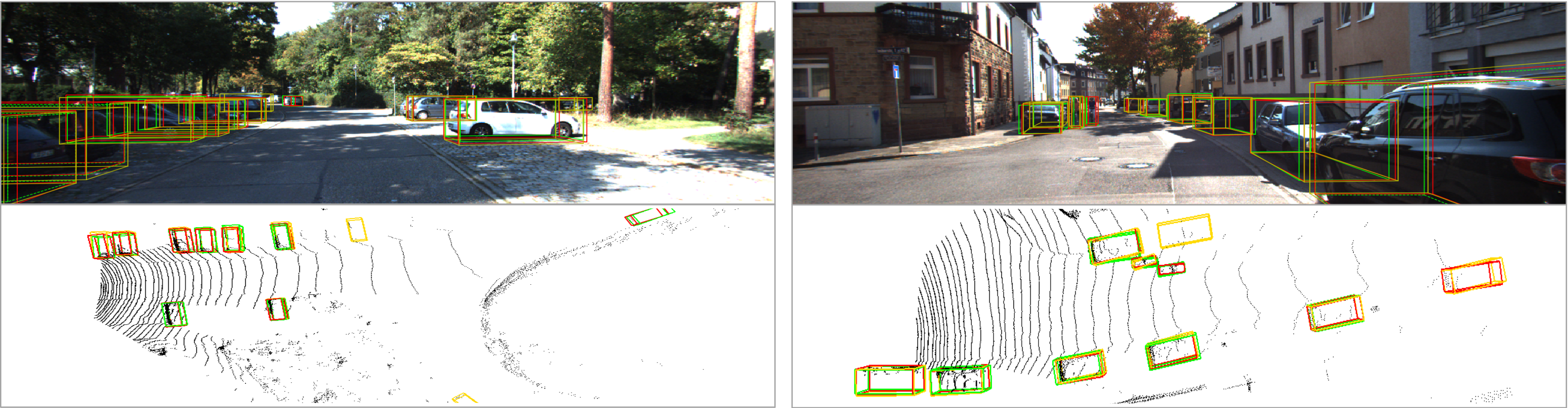}
 \end{center}
 \caption{Visualization of the results on the KITTI validation split set. The ground-truth 3D boxes, predicted 3D boxes of the baseline method, and predicted 3D boxes of our method are shown in green, yellow, and red, respectively, in the LiDAR phase. The first row and second row show the RGB images and the BEV images, respectively.}
 \label{fig6}
\end{figure*}

\noindent \textbf{The SC-reweight module for adaptation.}
\label{map}
The attention mechanism is widely employed in many vision tasks \cite{NonLocal2018, liu2019tanet:}. During adaptation, some parts of an object may substantially contribute to the 3D detection task and should be given greater attention, while other parts may have little contribution. We exploit a spatial and channel-wise loss reweighting module, which enables the framework to differentiate between distinct parts.

For spatial-wise attention, as shown in Figures \ref{fig5} and \ref{fig7}, the classification response heatmap is calculated by averaging the feature along the channel dimension. We can see that with a more complete point cloud structure, the object regions of the conceptual heatmap have higher values (warmer color) than the regions of the perceptual heatmap, which means that the higher response of the feature map reflects on the regions with more complete structure information. Based on this observation, we calculate the difference map between the two response maps to obtain an attention map for regions with added structural features. Consequently, the adaptation is strengthened in these regions and forces the network to be aware of invisible object regions. For channel-wise attention, the difference map between the conceptual and perceptual feature map is fed into a sub-module, named C-R, to obtain a \textbf{C}hannel-wise \textbf{R}eweighting vector. The C-R consists of an average pooling operation, several fully-connected layers, and a softmax activation. Next, the spatial reweighting map is tiled along the channel dimension to match the size of the original feature map. The reweighting vector is then multiplied by the spatial reweighting map to obtain the final reweighting map. 

This weighting approach shares the idea of focal loss \cite{lin2017focal}. We regard those parts of cars with fewer points and incomplete structure information as the hard-detected regions, which will be paid more attention during training and have much higher weights in loss computation. The SC-reweight is formulated as follows:
\begin{equation}
 \begin{split}
  \mathcal{M}_{reweight} =  \mathcal{\eta}\{ \mathcal{\phi}[(\frac{1}{J}\sum\limits_{j=1}^{J}{F_{class}^{\mathcal{P}}}^j - \frac{1}{J}\sum\limits_{j=1}^{J}{F_{class}^{\mathcal{C}}}^j)^2 \\
  \cdot \mathcal{M}_{FG}], J\} \cdot [1 + \mathcal{\beta}(F_{class}^{\mathcal{P}} - F_{class}^{\mathcal{C}})],
 \end{split}
 \label{eq6}
\end{equation}
where $\mathcal{M}_{reweight}$ is the output reweighting map of the SC-reweight, $F_{class}^{\mathcal{C}}$ and $F_{class}^{\mathcal{P}}$ are the inputs.
$J$ is the channels of $F_{class}$, $\mathcal{\phi}$ represents the operation that normalizes the map to 0$\sim$1. $\mathcal{\eta}\{ F, J \}$ denotes performing channel-wise tiling operation ($J$ times) on the feature $F$, and $\mathcal{M}_{FG}$ is the foreground mask.
$\mathcal{\beta}$ represents the feature mapping of the C-R.
$F_{class}^{\mathcal{P}}$ and $F_{class}^{\mathcal{C}}$ in Eq. \ref{eq6}, which are calculated for $\mathcal{M}_{reweight}$, are detached from the gradient descent.

\noindent \textbf{Biological model underlying the framework.}
\label{2}
The biological model of associative recognition, which underlies AGO-Net, includes three principal factors, as illustrated in Figure \ref{fig2}: visual cortex (VC), anterior temporal lobes (ATL), and inferior longitudinal fasciculus (ILF). The VC encodes the perceived primary information from real-world scenes, while the ATL have been implicated as a vital repository of conceptual knowledge in 3D object perception \cite{hoffman2018percept}. Different class-wise conceptual models stored in ATL are coded in an invariant manner with complete 3D structural features. ILF connects ATL to VC and provides the associative path between conceptual models and real-world objects. Weak features from occluded and distant objects are enhanced by the conceptual memory stored in the ATL. This conceptual knowledge retrieval and association process between the VC and the ATL is mimicked via domain adaptation.

\subsection{Training of AGO-Net}
\label{3}
\noindent \textbf{Training of the CFG.}
There are two losses for the CFG, including binary cross entropy loss for classification and smooth-$L1$ loss for proposal generation. We denote the total loss of the CFG as:
\begin{equation}
 \begin{array}{r}
  \mathcal{L}_{CFG}= \mathcal{L}_{box} + \mathcal{L}_{class}.
 \end{array}
 \label{eq1}
\end{equation}
Same regression targets as those in \cite{yan2018second, zhou2018voxelnet} are set up, and smooth-$L1$ loss $\mathcal{L}_{box}$ is adopted to regress the normalized box parameters as:
\begin{equation}
 \begin{array}{l}
  \Delta x = \frac{x_a-x_g}{d_a}, \Delta y = \frac{y_a-y_g}{h_a}, \Delta z = \frac{z_a-z_g}{d_a},    \\
  \Delta l = log(\frac{l_g}{l_a}), \Delta h = log(\frac{h_g}{h_a}), \Delta w = log(\frac{w_g}{w_a}), \\
  \Delta {\theta} = \theta_g - \theta_a,                                                             \\
 \end{array}
 \label{eq2}
\end{equation}
where $x$, $y$, and $z$ are the center coordinates; $w$, $l$, and $h$ are the width, length, and height, respectively; $\theta$ is the yaw rotation angle; the subscripts $a$, and $g$ indicate the anchor and the ground truth, respectively; and $d_a = \sqrt{(l_a)^2 + (w_a)^2}$ is the diagonal of the base of the anchor box. $(x_a, y_a, z_a, h_a, w_a, l_a, \theta_a)$ are the parameters of 3D anchors and $(x_g, y_g, z_g, h_g, w_g, l_g, \theta_g)$ represent the corresponding ground truth box.

We use the focal loss introduced by \cite{lin2017focal} for classification, to alleviate the sample imbalance during the anchor-based training, and the classification loss $\mathcal{L}_{class}$ is formulated as follows:
\begin{equation}
 \begin{array}{r}
  \mathcal{L}_{class}= \alpha_t(1-p_t)^\gamma log(p_t),
 \end{array}
 \label{eq3}
\end{equation}
where $p_t$ is the predicted classification probability and $\alpha$ and $\gamma$ are the parameters of the focal loss.

\noindent \textbf{Training of the entire AGO-Net.}
In addition to the classification and regression loss, there is an extra loss function for the feature adaptation. We denote the total loss of AGO-Net as $\mathcal{L}_{total}$:
\begin{equation}
 \begin{array}{r}
  \mathcal{L}_{total}= \mathcal{L}_{box} +  \mathcal{L}_{class} + \sigma \mathcal{L}_{ago},
 \end{array}
 \label{eq4}
\end{equation}
where $\sigma$ is a hyperparameter to balance loss terms. The association loss $\mathcal{L}_{ago}$ for feature adaptation is formulated as follows:
\begin{equation}
 \begin{split}
  \mathcal{L}_{ago}= \frac{1}{N} \sum_{n=1}^N  [\mathcal{S}_1 ({F_{box}^{\mathcal{P}}}^n, {F_{box}^{\mathcal{C}}}^n) \cdot (1 + {\mathcal{M}_{reweight}}^n) ],
 \end{split}
 \label{eq5}
\end{equation}
where $F_{box}^{\mathcal{P}}$ and $F_{box}^{\mathcal{C}}$ are the two feature maps for box regression from the perceptual (target) and conceptual (source) domains. $N$ denotes the number of nonzero pixels of $\mathcal{M}_{reweight}$ in Eq. \ref{eq6}, respectively. $\mathcal{S}_1$ denotes smooth-$L1$ distance.

\section{Experiments}
\label{sec:eval}
\subsection{Dataset and experimental setup}
We evaluate our method on three widely acknowledged 3D object detection datasets: KITTI \cite{geiger2012we}, nuScenes \cite{Caesar_2020_CVPR}, and Waymo Open Dataset (WOD) \cite{waymo}. 

\noindent \textbf{KITTI dataset.}
The KITTI dataset contains 7,481 training and 7,518 test images as well as the relevant point clouds. We split the training images equally into train and validation set following previous works such as \cite{yan2018second, Shi_2020_CVPR}. Average precision (AP) metrics measured in 3D and BEV are utilized to evaluate the performance. Three levels of difficulty are defined according to the 2D bounding box height, occlusion, and truncation degree as follows: ``Easy'', ``Mod.'', and ``Hard''. 

\noindent \textbf{nuScenes dataset.}
The nuScenes dataset contains 28,130 training samples, 6,019 validation samples, and 6,008 test samples. Each annotated frame corresponds to one point cloud captured by a 32-beam LiDAR. We follow the official evaluation protocol and report the AP (13 classes) and mean AP (mAP) results.

\noindent \textbf{Waymo dataset.}
Waymo Open Dataset includes RGB images and point clouds from five cameras and LiDAR sensors, respectively. It consists of 1,000 scenes for training and validation and 150 scenes for testing. The dataset has 12M labeled 3D objects which are annotated in a 360-degree field. For evaluation, AP and APH~\cite{waymo} are used as the metric.

\subsection{Implementation details}
We introduce the implementation details for the KITTI dataset in this section. More details of nuScenes and Waymo datasets can be found in the released code.

\noindent \textbf{Training details.}
For training, followed by \cite{yan2018second}, we select the LiDAR points of interest that lie in the range (0m, 70.4m), (-40m, 40m), (-3m, 1m) along X-, Y-, Z-axes, respectively, and discard the points that are invisible in the image view. During the training time for both CFG and PFE, we set the matching thresholds for the positive and negative anchors to 0.6 and 0.45, respectively. The matching IoU between the 3D bounding boxes and anchors is calculated by their nearest horizontal rectangles in BEV. The anchor for detecting the car has a size of 1.6m in width, 3.9m in length, and 1.56m in height. We discard all the anchors without any points. The construction of conceptual scenes is based only on the training set.

\begin{figure*}[htb]
 \begin{center}
  \includegraphics[width=16cm]{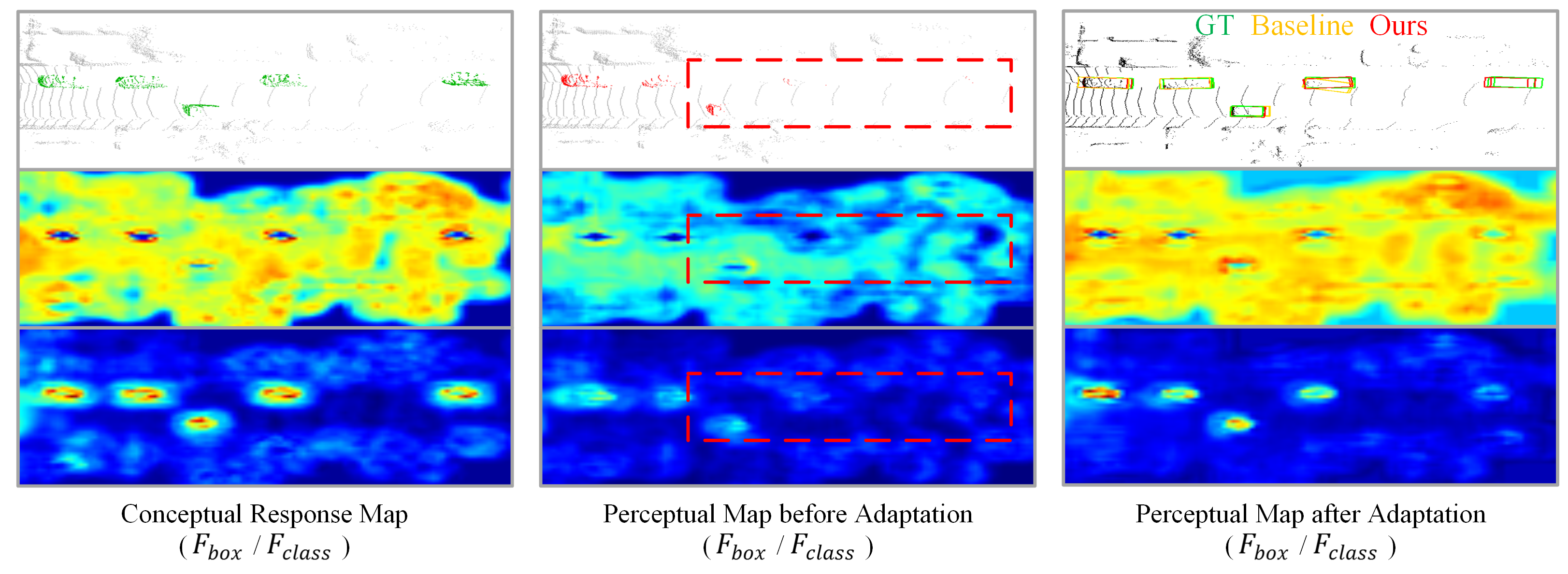}
 \end{center}
 \caption{Visualization of the feature maps and results on the KITTI validation set. The first row shows the BEV of the constructed conceptual scene and the perceptual scene, respectively. The ground-truth 3D boxes and the predicted 3D boxes of the baseline method and our method are drawn in green, yellow, and red, respectively, in the LiDAR phase.
  The second and third rows show the response maps of $F_{box}$ and $F_{class}$, respectively.}
 \label{fig7}
\end{figure*}

\noindent \textbf{Data augmentation.}
To prevent AGO-Net from over-fitting, we utilize the general ``cut-and-paste'' strategy following \cite{yan2018second}.
Note that for each training sample, we perform the same data augmentation for both conceptual and perceptual scenes so that the objects pasted in both scenes are spatially aligned for domain adaptation.

\noindent \textbf{Network architecture.}
Following \cite{He_2020_CVPR}, for the backbone network PFE, the point clouds of the scene are converted to the input tensor by quantization. After feeding into stacked sparse and sub-manifold convolutions for feature extraction and dimensionality reduction, the shape of the encoded tensor is 128$\times$2$\times$200$\times$176, where 2 corresponds to the height dimension. We squeeze the height dimension by reshaping it into the feature map channels. The box and classification head are then separately applied to decode the features for 3D object detection. Since the PFE and CFG are siamese structures with different input data, the architecture of CFG is identical to that of the PFE. We train the CFG for conceptual models end-to-end and then keep it fixed for further AGO-Net network training. In experiments, we utilize the SGD optimizer to train AGO-Net for 50 epochs on a single NVIDIA TITAN RTX GPU for about 14 hours. We set the learning rate, weight decay, and batch size of AGO-Net to 0.01, 0.001, and 2, respectively, and apply a cosine annealing strategy to decay the learning rate. For inference, the low-confidence bounding boxes are filtered out by a threshold of 0.3. We set the IoU threshold for non-maximum suppression (NMS) to 0.1.

\noindent \textbf{Training parameters.}
We set $\alpha=0.25$ and $\gamma=2$ in focal loss, $M=24$ and $K=20$ to create the conceptual scenes, $\sigma=1.0$ to balance the loss terms, and $\vartheta=0.25m$ to remove the added points that are close to the original points. For the point cloud quantization, we set $(S_x, S_y, S_z)$ as (0.05m, 0.05m, 0.1m).

\begin{table*}[htb]
 \caption{Results on the KITTI 3D object detection test server. The 3D object detection and BEV detection are evaluated by average precision at IoU = 0.7 for cars. According to the new policy of the KITTI dataset, all results are evaluated by the mean average precision with 40 recall positions (R40).}
 \small
 \centering
 \begin{tabular}{c|c|c|c|ccc|ccc|c}
  \toprule
  \hline
  \multirow{2} * {Stage}      & \multirow{2} * {Method}                  & \multirow{2} * {Time(s)} & \multirow{2} * {Modality} & \multicolumn{3}{c|}{3D Detection (\%)} & \multicolumn{3}{c|} {BEV Detection (\%)} & \multirow{2} * {GPU}                                                                 \\
                              &                                          &      &             & Easy           & Mod.           & Hard           & Easy           & Mod.           & Hard           &            \\
  \hline
  \multirow{11} * {Two-stage} & MV3D \cite{chen2017multi}                & 0.36 & LiDAR + RGB & 74.97          & 63.63          & 54.00          & 86.49          & 78.98          & 72.23          & TITAN X    \\
                              & F-PointNet \cite{qi2018frustum}          & 0.17 & LiDAR + RGB & 82.19          & 69.79          & 60.59          & 91.17          & 84.67          & 74.77          & GTX 1080   \\
                              & AVOD     \cite{ku2018joint}              & 0.10 & LiDAR + RGB & 76.39          & 66.47          & 60.23          & 89.75          & 84.95          & 78.32          & TITAN XP   \\
                              & PointRCNN    \cite{shi2019pointrcnn}     & 0.10 & LiDAR only  & 86.96          & 75.64          & 70.70          & 92.13          & 87.39          & 82.72          & TITAN XP   \\
                              & F-ConvNet\cite{wang2019frustum}          & 0.48 & LiDAR + RGB & 87.36          & 76.39          & 66.69          & 91.51          & 85.84          & 76.11          & -          \\
                              & Fast Point R-CNN    \cite{chen2019fast}  & 0.07 & LiDAR only  & 85.29          & 77.40          & 70.24          & 90.87          & 87.84          & 80.52          & Tesla P40  \\
                              & MMF \cite{liang2019multi}                & 0.08 & LiDAR + RGB & 88.40          & 77.43          & 70.22          & 93.67          & 88.21          & 81.99          & TITAN XP   \\
                              & STD \cite{yang2019std}                   & 0.10 & LiDAR only  & 87.95          & 79.71          & 75.09          & 94.74          & 89.19          & 86.42          & TITAN V    \\
                              & Patches \cite{lehner2019patch}           & 0.15 & LiDAR only  & 88.67          & 77.20          & 71.82          & 92.72          & 88.39          & 83.19          & GTX 1080Ti \\
                              & Part-$A^2$ \cite{shi2020from}            & 0.07 & LiDAR only  & 87.81          & 78.49          & 73.51          & 91.70          & 87.79          & 84.61          & Tesla V100 \\
                              & PV-RCNN     \cite{Shi_2020_CVPR}         & 0.15 & LiDAR only  & 90.25          & \textbf{81.43} & \textbf{76.82} & 94.98          & 90.65          & \textbf{86.14} & GTX 2080Ti \\
  \hline
  \multirow{8} * {One-stage}  & VoxelNet \cite{zhou2018voxelnet}         & 0.23 & LiDAR only  & 77.82          & 64.17          & 57.51          & 87.95          & 78.39          & 71.29          & TITAN X    \\
                              & ContFuse \cite{liang2018deep}            & 0.06 & LiDAR + RGB & 83.68          & 68.78          & 61.67          & 94.07          & 85.35          & 75.88          & -          \\
                              & SECOND    \cite{yan2018second}           & 0.05 & LiDAR only  & 83.34          & 72.55          & 65.82          & 89.39          & 83.77          & 78.59          & GTX 1080Ti \\
                              & PointPillars \cite{lang2019pointpillars} & 0.02 & LiDAR only  & 82.58          & 74.31          & 68.99          & 90.07          & 86.56          & 82.81          & GTX 1080Ti \\
                              & Point-GNN \cite{Shig_2020_CVPR}          & 0.64 & LiDAR only  & 88.33          & 79.47          & 72.29          & 93.11          & 89.17          & 83.90          & GTX 1070   \\
                              & 3DSSD \cite{Yang_2020_CVPR}              & 0.04 & LiDAR only  & 88.36          & 79.57          & 74.55          & 92.66          & 89.02          & 85.86          & TITAN V    \\
                              & SA-SSD     \cite{He_2020_CVPR}           & 0.04 & LiDAR only  & 88.75          & 79.79          & 74.16          & 95.03          & \textbf{91.03} & 85.96          & GTX 2080Ti \\
                              & AGO-Net (ours)                           & 0.04 & LiDAR only  & \textbf{91.53} & 80.77          & 75.23          & \textbf{95.55} & 90.00          & 84.72          & GTX 2080Ti \\
  \hline
 \end{tabular}
 \label{tab1}
\end{table*}

\subsection{Quantitative results}
\noindent \textbf{KITTI dataset.} Table \ref{tab1} shows the results on the KITTI 3D object detection test server. On 08.10.2019, KITTI changed its evaluation setting by using 40 recall positions. We compare the proposed AGO-Net with other SoTA approaches by submitting the detection results to the KITTI server for evaluation. At the time of submission (22 Jun. 2020), our AGO-Net ranked $1^{st}$ among all existing one-stage detectors, $2^{nd}$ among all existing detectors, and $1^{st}$ on the ``Easy'' entry among all competitors on the leaderboard in the most important ``Car'' category, with a real-time inference speed of 25 FPS (faster than almost all other 3D object detection methods). Our method is three times faster than the two-stage top-ranked method PV-RCNN \cite{Shi_2020_CVPR}, while our ``Easy'' entry is even better than PV-RCNN, and the ``Mod.'' and ``Hard'' entries are close to it. Superior performance can be observed from Table \ref{tab1} when comparing AGO-Net with other cutting-edge one-stage 3D detection approaches. Our AGO-Net leads both SA-SSD \cite{He_2020_CVPR} and 3DSSD \cite{Yang_2020_CVPR} by (2.8\%; 1.0\%; 1.1\%) and (3.2\%; 1.2\%; 0.7\%) in 3D detection, respectively. The performance boost is mainly contributed by the domain adaptation-based feature association, which exploits the feature enhancement capability of the network.

\begin{table}[htb]
 \caption{ ``Mod.'' results on 3D object detection of the KITTI validation split set at IoU = 0.7 for cars. The AP is calculated with 11 recall positions (R11) to compare our method with the previous methods.}
 \small
 \centering
 \begin{tabular}{c|C{2.5cm}|C{2cm}|C{1.2cm}}
  \toprule
  \hline
  Stage                 & Method                                  & Reference    & $\rm AP_{3D}$      \\
  \hline
  \multirow{4} * {Two-stage}
                        & STD \cite{yang2019std}                  & ICCV 2019    & 79.80          \\
                        & Patches \cite{lehner2019patch}          & arXiv 2019   & 79.04          \\
                        & Part-$A^2$ \cite{shi2020from}           & TPAMI 2020   & 79.47          \\
                        & PV-RCNN     \cite{Shi_2020_CVPR}        & CVPR 2020    & 83.90          \\
  \hline
  \multirow{4} * {One-stage}
                        & Point-GNN \cite{Shig_2020_CVPR}         & CVPR 2020    & 78.34          \\
                        & 3DSSD \cite{Yang_2020_CVPR}             & CVPR 2020    & 79.45          \\
                        & SA-SSD  \cite{He_2020_CVPR}             & CVPR 2020    & 79.91          \\
                        & AGO-Net (ours)                          & -            & \textbf{83.92} \\
  \hline
 \end{tabular}
 \label{tab2}
\end{table}

To further validate the superior performance of AGO-Net, we evaluate AGO-Net on the KITTI validation split set, as shown in Table \ref{tab2}. Note that we only report AP with 11 recall positions in Table \ref{tab2} and Table \ref{tab12} to compare with the results from previous frameworks. For 3D object detection, by only utilizing LiDAR point clouds, our AGO-Net outperforms all existing methods on the most critical ``Mod.'' difficulty level. Thanks to the self-contained conceptual model and feature association mechanism, our straightforward yet effective network can achieve superior performance compared with complicated networks in both accuracy and speed. AGO-Net leads the $1^{st}$ ranked one-stage method SA-SSD and current $1^{st}$ ranked two-stage method PV-RCNN by 4.0\% and 0.2\%, respectively.

\begin{table}[htb]
 \caption{3D object detection $\rm AP_{3D}$ (R40) performance for ``Pedestrian'' and ``Cyclist'' on KITTI val split set at IoU = 0.5.}
 \small
 \centering
 \begin{tabular}{C{2.1cm}|C{0.63cm}C{0.63cm}C{0.63cm}|C{0.63cm}C{0.63cm}C{0.63cm}}
  \toprule
  \hline
  \multirow{2} * {Method}         & \multicolumn{3}{c|}{Pedestrian}                    & \multicolumn{3}{c} {Cyclist}                  \\
                          & Easy            & Mod.            & Hard            & Easy            & Mod.            & Hard             \\
  \hline
  conceptual              & 61.79           & 58.77           & 59.97           & 84.99           & 80.87           & 80.56            \\
  \hline
  SECOND$\rm^{\dag}$ \cite{yan2018second}                 & 55.94           & 51.14           & 46.17           & 82.96           & 66.74           & 62.78             \\
  baseline                & 57.47           & 52.02           & 47.44           & 83.31           & 67.55           & 63.94             \\
  ours                 & \textbf{60.39}  & \textbf{54.81}  & \textbf{50.59}  & \textbf{87.57}  & \textbf{69.24}  & \textbf{64.74}    \\
  \hline
 \end{tabular}
 \label{tab3}
\end{table}

\begin{table}[htb]
 \caption{3D object detection $\rm AP_{3D}$ (R40) performance for ``Pedestrian'' and ``Cyclist'' on KITTI test set at IoU = 0.5.}
 \small
 \centering
 \begin{tabular}{C{2.1cm}|C{0.63cm}C{0.63cm}C{0.63cm}|C{0.63cm}C{0.63cm}C{0.63cm}}
  \toprule
  \hline
  \multirow{2} * {Method}         & \multicolumn{3}{c|}{Pedestrian}                    & \multicolumn{3}{c} {Cyclist}                  \\
                          & Easy            & Mod.            & Hard            & Easy            & Mod.            & Hard             \\
  \hline
  SECOND$\rm^{\dag}$ \cite{yan2018second}& 43.04      & 35.92      & 33.55      & 71.05           & 55.64           & 49.83             \\
  ours                                   & \textbf{45.18}  & \textbf{37.22}  & \textbf{34.62}  & \textbf{72.82}  & \textbf{57.60}  & \textbf{51.53}    \\
  \hline
 \end{tabular}
 \label{tab4}
\end{table}

The ``Cyclist'' and ``Pedestrian'' categories are more challenging than the ``Car'' for voxel-only-based 3D object detection due to their non-rigid structures and small scale. To fully validate the effectiveness of the proposed method, we still report $\rm AP_{3D}$ results (IoU = 0.5) of these two categories with respect to the baseline method and SECOND \cite{yan2018second} in Tables \ref{tab3} and \ref{tab4}. The SECOND$\rm^{\dag}$ is implemented in OpenPCDet~\cite{openpcdet2020}. Note that in the following sections, the ``baseline'' denotes our instantiated PFE (without the CFG) introduced in Sec. \ref{1}, and the ``conceptual'' denotes training and testing on the conceptual dataset. Despite the number and quality limitation of conceptual models, our method still outperforms the baseline method with considerable margins.

As shown in Table \ref{tab8}, we also report results for the near-range experiment at 0-30 m, median-range at 30-50 m, and far-range at 50-80 m with and without adaptation. Regarding both near-, median-, and far-range performances, our method outperforms the baseline method by (0.5\%; 3.1\%; 3.0\%) on ``Mod.'' entry, respectively, especially for the median- and far-range.

\noindent \textbf{nuScenes and Waymo datasets.}
To demonstrate the versatility of our approach, we evaluate on nuScenes and Waymo datasets. Following previous works \cite{lang2019pointpillars,Shi_2020_CVPR}, we adopt the same range and ring-view scenes for processing point clouds and official evaluation metrics of each dataset. Tables \ref{tab5} and \ref{tab6} demonstrate the superiority of our method on validation and test set of nuScenes, and Table \ref{tab7} depicts the prevalence on the Waymo dataset compared to the baseline and the cutting-edge one-stage method \cite{yan2018second}. For Waymo dataset, only 20\% subset of data is utilized for training due to limited computation resources, which is consistent with OpenPCDet~\cite{openpcdet2020}. Note that the backbone of our method (derived from \cite{He_2020_CVPR}) is different from SECOND$\rm^{\dag}$, therefore the performances of SECOND$\rm^{\dag}$ and the baseline are slightly different.

\begin{table*}[htb]
 \caption{
  Per class performance on the nuScenes validation set. Evaluation of detections as measured by average precision (AP) or mean AP (mAP). Abbreviations: construction vehicle (C. V.), pedestrian (Ped.), motorcycle (Motor.), and traffic cone (T. C.).
 }
 \small
 \centering
 \begin{tabular}{c|c|c|c|c|c|c|c|c|c|c|c|c}
  \toprule
  \hline
  Methods      & mAP           & Car           & Truck         & Bus           & Trailer       & C. V.     & Ped.          & Motor.    & Bicycle      & T. C.      & Barrier       & Reference     \\
  \hline
  SECOND \cite{yan2018second}       & 27.1          & 75.5          & 21.9          & 29.0          & 13.0          & 0.4           & 59.9          & 16.9          & 0.0          & 22.5          & 32.2          & Sensors 2018  \\
  PointPillars \cite{lang2019pointpillars} & 29.5          & 70.5          & 25.0          & 34.4          & 20.0          & 4.5           & 59.9          & 16.7          & 1.6          & 29.6          & 33.2          & CVPR 2019     \\
  WYSIWYG \cite{Hu_2020_CVPR}     & 35.4          & 80.0          & 35.8          & 54.1          & 28.5          & 7.5           & 66.9          & 18.5          & 0.0          & 27.9          & 34.5          & CVPR 2020     \\
  InfoFocus \cite{InfoFocus}   & 36.4          & 77.6          & 35.4          & 50.5          & 25.6          & 8.3           & 61.7          & 25.2          & 2.5          & 33.4          & 43.4          & ECCV 2020     \\
  3DSSD \cite{Yang_2020_CVPR}       & 42.7          & 81.2          & 47.2          & 61.4          & 30.5          & 12.6          & 70.2          & \textbf{36.0}          & \textbf{8.6}          & 31.1          & 47.9          & CVPR 2020     \\
  \hline
  baseline     & 42.9          & 79.8          & 45.9          & 62.0          & 32.5          & 12.7          & 71.1          & 29.6          & 4.5          & 44.1          & 47.2          & -             \\
  ours         & \textbf{45.1} & \textbf{81.5} & \textbf{50.1} & \textbf{62.2} & \textbf{34.0} & \textbf{13.3} & \textbf{72.2} & 32.5 & 5.9 & \textbf{48.1} & \textbf{51.2} & -             \\
  \hline
 \end{tabular}
 \label{tab5}
\end{table*}

\begin{table*}[htb]
 \caption{
  Per class performance on the nuScenes test set. Evaluation of detections as measured by average precision (AP) or mean AP (mAP). Abbreviations: construction vehicle (C. V.), pedestrian (Ped.), motorcycle (Motor.), and traffic cone (T. C.).
 }
 \small
 \centering
 \begin{tabular}{c|c|c|c|c|c|c|c|c|c|c|c|c}
  \toprule
  \hline
  Methods      & mAP  & Car  & Truck & Bus  & Trailer & C. V. & Ped. & Motor. & Bicycle & T. C. & Barrier & Reference \\
  \hline
  PointPillars \cite{lang2019pointpillars}& 30.5 & 68.4 & 23.0  & 28.2 & 23.4    & 4.1       & 59.7 & 27.4       & 1.1     & 30.8     & 38.9    & CVPR 2019 \\
  CenterNet \cite{centernet}   & 33.8 & 53.6 & 27.0  & 24.8 & 25.1    & 8.6       & 37.5 & 29.1       & \textbf{20.7}    & \textbf{58.3}     & 53.3    & arXiv 2019 \\
  WYSIWYG \cite{Hu_2020_CVPR}     & 35.0 & 79.1 & 30.4  & 46.6 & 40.1    & 7.1       & 65.0 & 18.2       & 0.1     & 28.8     & 34.7    & CVPR 2020 \\
  InfoFocus \cite{InfoFocus}   & 39.5 & 77.9 & 31.4  & 44.8 & 37.3    & 10.7      & 63.4 & 29.0       & 6.1     & 46.5     & 47.8    & ECCV 2020 \\
  \hline
  ours         & \textbf{46.2} & \textbf{80.7} & \textbf{45.0}      & \textbf{57.5}      & \textbf{48.4}       & \textbf{12.0}          & \textbf{71.4}     & \textbf{30.7}           & 5.9        & 57.2         & \textbf{53.6}       &- \\
  \hline
 \end{tabular}
 \label{tab6}
\end{table*}

\begin{table*}[htb]
 \caption{
  Multi-class 3D detection results on the validation sequence of the Waymo dataset.
 }
 \small
 \centering
 \begin{tabular}{C{2.0cm}|C{2.6cm}|C{1.55cm}C{1.55cm}|C{1.55cm}C{1.55cm}|C{1.55cm}C{1.55cm}}
  \toprule
  \hline
  \multirow{2} * {Difficulty} & \multirow{2} * {Method} 
  & \multicolumn{2}{c|}{Vehicle (IoU=0.7)} & \multicolumn{2}{c|}{Pedestrian (IoU=0.5)} & \multicolumn{2}{c}{Cyclist (IoU=0.5)}   \\
  &                                                   
  & 3D AP             & 3D APH             & 3D AP             & 3D APH                & 3D AP           & 3D APH                     \\
  \hline
  \multirow{3} * {LEVEL\_1}  & SECOND$\rm^{\dag}$ \cite{yan2018second}  & 67.9 & 67.3 & 57.8 & 47.6 & 54.0 & 52.7 \\
                            & baseline                     & 67.4 & 66.8 & 57.4 & 47.8 & 53.5 & 52.3 \\
                            & ours                         &\textbf{69.2}&\textbf{68.7} & \textbf{59.3} & \textbf{48.7} & \textbf{55.3} & \textbf{54.2} \\
  \hline
  \multirow{3} * {LEVEL\_2}  & SECOND$\rm^{\dag}$ \cite{yan2018second}  & 59.4 & 58.9 & 49.8 & 41.0 & 52.3 & 51.0 \\
                            & baseline                     & 58.9 & 58.3 & 49.4 & 41.1 & 51.8 & 50.6 \\
                            & ours                         &\textbf{60.6}&\textbf{60.1} & \textbf{51.8} & \textbf{42.4} & \textbf{53.5} &\textbf{52.5}\\
  \hline
 \end{tabular}
 \label{tab7}
\end{table*}

\begin{table}[htb]
 \caption{Near-range (0$\sim$30m), median-range (30$\sim$50m), and far-range (50$\sim$80m) comparisons on $\rm AP_{3D}$ (R40). Note that for the easy level in the far-range area, the AP is always 0 since there is no easy object.}
 \small
 \centering
 \begin{tabular}{C{0.9cm}|C{0.43cm}C{0.45cm}C{0.63cm}|C{0.43cm}C{0.45cm}C{0.63cm}|C{0.43cm}C{0.63cm}}
  \toprule
  \hline
  \multirow{2} * {Method} & \multicolumn{3}{c|}{Near-range} & \multicolumn{3}{c|} {Median-range} & \multicolumn{2}{c} {Far-range}            \\
           & Easy           & Mod.           & Hard           & Easy           & Mod.           & Hard           & Mod.           & Hard           \\
  \hline
  baseline & 95.37          & 95.33          & 92.34          & 48.05          & 63.89          & 61.73          & 7.49           & 9.57           \\
  ours     & \textbf{95.91} & \textbf{95.85} & \textbf{92.90} & \textbf{52.14} & \textbf{67.02} & \textbf{64.79} & \textbf{10.44} & \textbf{11.36} \\
  \hline
 \end{tabular}
 \label{tab8}
\end{table}

\subsection{Qualitative results}
We present some representative comparison results of the baseline and AGO-Net on the KITTI validation set in Figure \ref{fig6}. 3D bounding boxes detected from LiDAR are projected onto RGB images for better visualization. The ground truth, and the predicted 3D boxes of the baseline and our method, are colored in green, yellow, and red, respectively. The first to third rows show the RGB image, the front view, and the bird’s-eye view, respectively. As observed, our method can produce high-quality 3D bounding boxes in different kinds of scenes. The distant and occluded objects that are difficult for the baseline method to detect, are accurately predicted by AGO-Net due to the association capability of the network.

To further observe how the perceptual-to-conceptual domain adaptation fundamentally improves 3D object detection, we visualize the conceptual and  perceptual features before and after adaptation as well as their corresponding detection results in Figure \ref{fig7}. The first row shows the bird’s-eye view of perceptual and conceptual scenes, respectively. The second and third rows show the feature decoded for 3D bounding box and classification, respectively. We average the feature map along the channel dimension to obtain the response heatmap. Warmer color indicates a higher response. As depicted in the first and second columns, the pixel-wise difference between two feature maps is prone to be large in regions containing insufficient and incomplete structural information due to occlusion or distance. The conceptual response of distant and occluded objects is much more obvious than that of the perceptual objects (highlighted by the red boxes), as the added point clouds for each incomplete perceptual object provide adequate structural information to extract discriminative features. After the perceptual domain feature is well adapted, the response of the distant and occluded objects becomes more similar to that of conceptual objects. The qualitative comparison (Row 1 and Column 3) of the results based on the original perceptual feature (baseline) and the feature after adapted (ours) shows that the two distant objects are well detected, and the rotation angle of the object is refined due to the associated critical structural information. The feature visualization demonstrates that the proposed approach adaptively generates robust features for predicting more accurate 3D bounding boxes.

\subsection{Ablation study}
Ablation experiments are conducted on the KITTI dataset. 

\begin{table}[htb]
 \caption{The upper bound performance of $\rm AP_{3D}$ (R40) training with the conceptual data and testing on the conceptual.}
 \small
 \centering
 \begin{tabular}{c|c|C{0.85cm}C{0.85cm}C{0.85cm}}
  \toprule
  \hline
  \multirow{2} * {training data} & \multirow{2} * {validation data} & \multicolumn{3}{c}{$\rm AP_{3D}$ (IoU=0.7)}                 \\
                                 &                       & Easy               & Mod.               & Hard                         \\
  \hline
  conceptual                     & conceptual            & 98.74              & 98.72              & 96.56                        \\
  \hline
  conceptual                     & real                  & 92.39              & 72.55              & 70.04                        \\
  real                           & real                  & 92.21              & \textbf{83.03}     & \textbf{79.98}               \\
  real + conceptual              & real                  & \textbf{92.42}     & 82.97              & 79.86                        \\
  \hline
 \end{tabular}
 \label{tab9}
\end{table}

\noindent \textbf{The upper bound performance of the conceptual domain.}
To validate the upper-bound performance of the CFG module, we conduct experiments on conceptual scenes only in both training and testing, rather than the real dataset. High performance can be observed in the first row of Table \ref{tab9}, which can serve to instruct our associated-guided network. As shown in the first row, the high precision (over 96\%) on all entries indicates the possibility of adopting conceptual models to guide PFE for enhanced feature learning, while the second row shows the domain gap between the conceptual and perceptual dataset; therefore we cannot just use the conceptual scenes for training and test the model on the real scenes. The third row shows the performance of the baseline model. The fourth row ``real + conceptual'' indicates merely training baseline model with KITTI train split set and the generated conceptual data without using the siamese network for adaptation. The result validates that simply mixing more training data using the same network will not improve the detector performance.

\begin{table}[htb]
 \caption{AGO-Net with different settings on $\rm AP_{3D}$ at IoU=0.7 (R40).}
 \small
 \centering
 \begin{tabular}{C{3.3cm}|C{1.2cm}C{1.2cm}C{1.2cm}}
  \toprule
  \hline
  \multirow{2} * {Method}                    & \multicolumn{3}{c}{$\rm AP_{3D}$ (IoU=0.7)}        \\
                                             & Easy           & Mod.           & Hard    \\
  \hline
  baseline                                   & 92.21          & 83.03          & 79.98    \\
  \hline
  CFG                                        & 93.45          & 84.21          & 81.86     \\
  CFG + SC                          & \textbf{94.01} & \textbf{84.84} & \textbf{82.38}      \\
  \hline
 \end{tabular}
 \label{tab10}
\end{table}


\noindent \textbf{Different settings of AGO-Net.}
We investigated the performance with and without the domain adaptation and the SC-reweight module, as shown in Table \ref{tab10}. The first row is the baseline results trained with only KITTI train split set. Owing to the object-wise domain adaptation (``CFG''), an improvement is observed in Row 2. The third row shows the results with both the CFG and the SC-reweight module (``SC''). The improvements on all difficulty levels indicate that the full approach learns more discriminative features for 3D detection.

\begin{table}[htb]
 \caption{Effects of spatial and channel-wise attention in SC-reweight (R40).}
 \small
 \centering
 \begin{tabular}{C{3.3cm}|C{1.2cm}C{1.2cm}C{1.2cm}}
  \toprule
  \hline
  \multirow{2} * {Attention setting} & \multicolumn{3}{c}{$\rm AP_{3D}$ (IoU=0.7)}                                   \\
                    & Easy           & Mod.           & Hard           \\
  \hline
  none              & 93.45          & 84.21          & 81.86          \\
  spatial           & 93.90          & 84.75          & 82.32          \\
  channel           & 93.59          & 84.32          & 81.97          \\
  spatial + channel & \textbf{94.01} & \textbf{84.84} & \textbf{82.38} \\
  \hline
 \end{tabular}
 \label{tab11}
\end{table}

\noindent \textbf{Spatial and channel-wise reweighting (SC-reweight).}
To validate the effectiveness of the SC-reweight, we conduct experiments with different settings of the SC-reweight, as shown in Table \ref{tab11}. ``spatial'' shows the performance based on the reweighting map that only utilizes spatial attention. ``channel'' reports the result with channel-wise attention. ``spatial + channel'' shows the performance with the full setting. The progressive improvement demonstrates that the spatial and channel-wise loss reweighting is vital to the domain adaptation process and forces the network to adapt to more critical structural features during the adaptation.

\begin{figure}[htb]
 \begin{center}
  \includegraphics[width=8.2 cm]{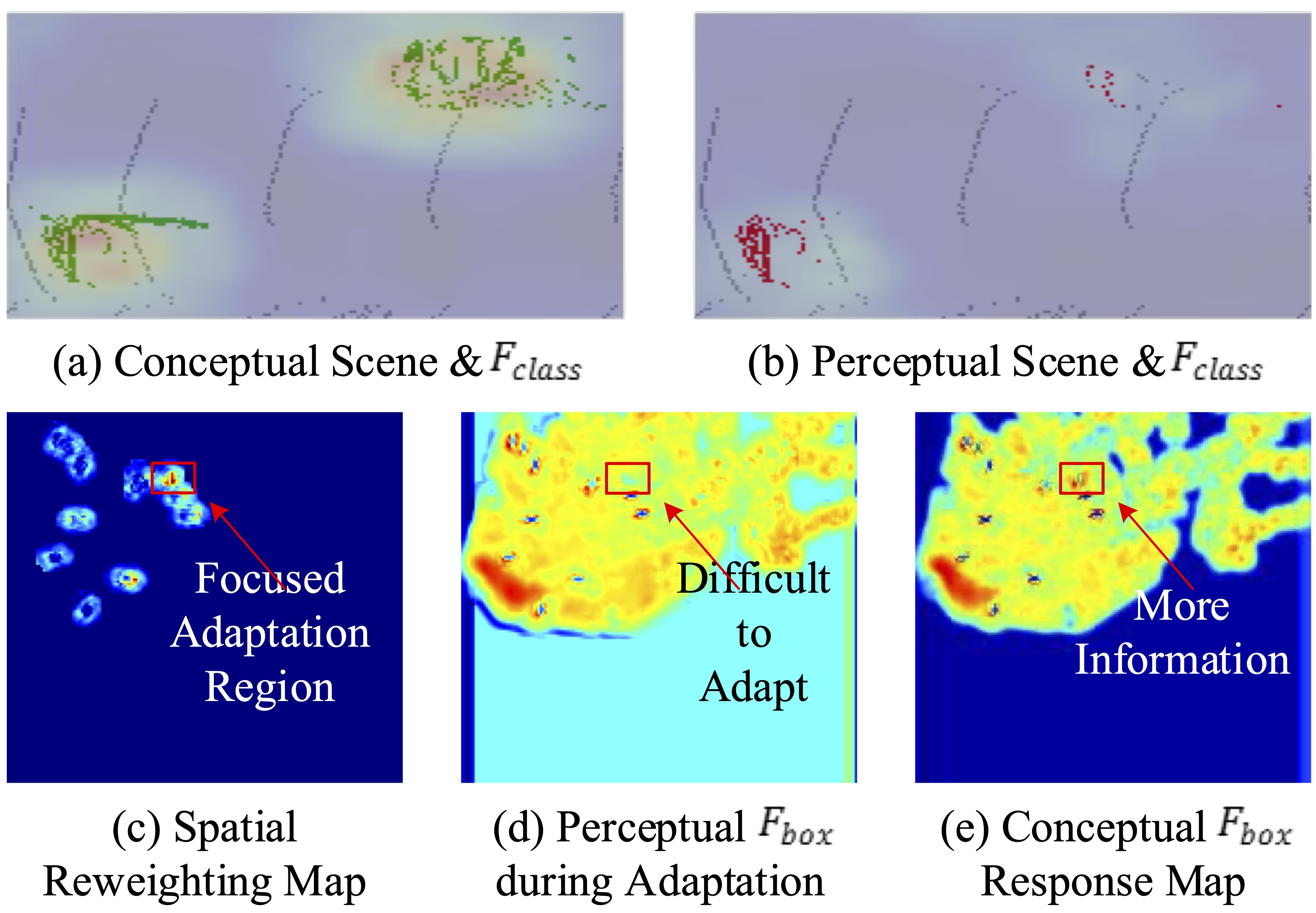}
 \end{center}
 \caption{Visualization of the feature maps and spatial reweighting maps. The first row shows the comparison of (a) conceptual scene and (b) perceptual scene with their classification response maps, respectively. The second row depicts (c) the spatial reweighting map, (d) the perceptual feature response map during adaptation, and (e) the conceptual feature response map, respectively.}
 \label{fig8}
\end{figure}

The comparison of (a) and (b) in Figure \ref{fig8} shows that the regions with added point clouds in the conceptual scene have a much higher response than the regions in the perceptual scene. As shown in (c), the high-response regions of the spatial reweighting map are spatially aligned with those less informative regions that are difficult to adapt in the perceptual feature in (d). Consequently, during the adaptation process, the regions are given more attention to adapt more spatial and structural information for regressing 3D object bounding boxes from the conceptual feature in (e).

\begin{table}[htb]
 \caption{Applying our association-guided architecture on different frameworks on $\rm AP_{3D}$. The AP is calculated with 11 recall positions (R11) to compare our method with the previous methods}
 \small
 \centering
 \begin{tabular}{C{3.3cm}|C{1.2cm}C{1.2cm}C{1.2cm}}
  \toprule
  \hline
  \multirow{2} * {Method}                  & \multicolumn{3}{c}{$\rm AP_{3D}$ (IoU=0.7)}      \\
                                           & Easy           & Mod.           & Hard           \\
  \hline
  PointPillars \cite{lang2019pointpillars} & 85.14          & 76.44          & 70.02          \\
  PointPillars + AGO                       & \textbf{86.16} & \textbf{78.53} & \textbf{75.21} \\
  \hline
  SECOND \cite{yan2018second}              & 88.07          & 77.12          & 75.27          \\
  SECOND + AGO                             & \textbf{89.17} & \textbf{78.82} & \textbf{77.51} \\
  \hline
 \end{tabular}
 \label{tab12}
\end{table}

\noindent \textbf{Generalizability.} 
We adopt other 3D detection networks \cite{lang2019pointpillars, yan2018second} to instantiate the PFE and CFG to validate the expansion capability and versatility of AGO-Net. As shown in Table \ref{tab12}, our association-guided framework (with ``AGO'') boosts these methods by a large margin, especially for ``Mod.'' and ``Hard'' entries.

\begin{table}[htb]
 \caption{Comparison on different construction strategies (R40).}
 \small
 \centering
 \begin{tabular}{C{3.3cm}|C{1.2cm}C{1.2cm}C{1.2cm}}
  \toprule
  \hline
  \multirow{2} * {Construction strategy} & \multicolumn{3}{c}{$\rm AP_{3D}$ (IoU=0.7)}                                   \\
                     & Easy           & Mod.           & Hard           \\
  \hline
  replace            & 93.74          & 84.38          & 82.11          \\
  add (with removal) & \textbf{94.01} & \textbf{84.84} & \textbf{82.38} \\
  \hline
 \end{tabular}
 \label{tab13}
\end{table}

\noindent \textbf{Different conceptual scene construction strategies.}
As illustrated in Figure \ref{fig4}, different from the previous work \cite{Du_2020_CVPR}, we do not simply use the conceptual model to replace the perceptual object. Instead, we paste the conceptual model onto the original perceptual object and remove those conceptual points that are close to the original perceptual points. Such a simple removal operation keeps the original structure of the object point cloud.
Table \ref{tab13} shows the superior performance of our improved strategy.

\begin{table}[htb]
 \caption{Comparison of performing adaptation on different features (R40).}
 \small
 \centering
 \begin{tabular}{C{3.3cm}|C{1.2cm}C{1.2cm}C{1.2cm}}
  \toprule
  \hline
  \multirow{2} * {Feature for adaptation} & \multicolumn{3}{c}{$\rm AP_{3D}$ (IoU=0.7)}                                   \\
                      & Easy           & Mod.           & Hard           \\
  \hline
  $F_{box}+F_{class}$ & 93.40          & 84.10          & 81.73          \\
  $F_{class}$         & 92.51          & 83.16          & 80.07          \\
  $F_{box}$           & \textbf{93.45} & \textbf{84.21} & \textbf{81.86} \\
  \hline
 \end{tabular}
 \label{tab14}
\end{table}

\noindent \textbf{Comparison of performing adaptation on different features.}
We attempt to adapt only $F_{class}$ or both $F_{box}$ and $F_{class}$ to ascertain which feature adaptation is more effective. As shown in Table \ref{tab14}, ``$F_{box} + F_{class}$'' adaptation and ``$F_{class}$'' adaptation do not perform better than ``$F_{box}$'' adaptation. The association-based mechanism digs deep into the invisible structures of occluded and distant objects for object perception, with the premise that the categories of the objects are correctly recognized.

\noindent \textbf{Comparison on different attention modules.}
We also experiment to compare the SC-reweight module with the Incompletion-aware module (IAM) proposed in \cite{Du_2020_CVPR}. Table \ref{tab15} demonstrates the superior performance of the SC-reweight module. Note that the SC-reweight does not depend on additional time-consuming deformable convolutional layers \cite{dai2017deformable}.

\begin{table}[h]
 \caption{Comparison on different attention modules with and without the deformable convolutional layer (R40).}
 \small
 \centering
 \begin{tabular}{C{1.8cm}|C{1cm}|C{1cm}C{1cm}C{1cm}|C{0.5cm}}
  \toprule
  \hline
  Attention           & Deform            & \multicolumn{3}{c|}{$\rm AP_{3d}$ (IoU=0.7)}                      & \multirow{2} * {FPS}          \\
  module              & layer                 & Easy                & Mod.                 & Hard                 &                               \\
  \hline
  none                & $\times$              & 93.45               & 84.21                & 81.86                & 25                            \\
  none                & $\checkmark$          & 93.71               & 84.39                & 82.02                & 17                            \\
  IAM~\cite{Du_2020_CVPR} & $\checkmark$          & 94.08               & 84.69                & 82.37                & 17                            \\
  SC-reweight         & $\times$              & 94.01               & 84.84                & 82.38                & 25                            \\
  SC-reweight         & $\checkmark$          & \textbf{94.26}      & \textbf{84.98}       & \textbf{82.52}       & 17                            \\
  \hline
 \end{tabular}
 \label{tab15}
\end{table}

\section{Conclusion}
\label{sec:conclusion}
In this paper, we proposed a brain-inspired 3D point cloud object detection framework, named AGO-Net. This learns to associate features of perceived objects from the real scene with discriminative features from their conceptual models via domain adaptation, which fundamentally enhances the feature robustness against appearance changes in point clouds. We further leverage the classification feature maps to obtain a reweighting map for the adaptation loss, which encourages the feature adaptation to focus on more informative regions, and boosts the final detection performance. A practical method for generating conceptual scenes without any external dataset is also introduced. Moreover, our simple yet effective approach can be easily integrated into many existing 3D object detection frameworks. The experimental results on the KITTI, nuScenes and Waymo datasets demonstrate the effectiveness and efficiency of the proposed framework.

\ifCLASSOPTIONcompsoc
\section*{Acknowledgments}
This work was supported by 
National Key R\&D Program of China (No.2019YFA0709502, 2018YFC1312904),
Shanghai Municipal Science and Technology Major Project (No.2018SHZDZX01), ZJ Lab, and Shanghai Center for Brain Science and Brain-Inspired Technology, the 111 Project (No.B18015).

\begin{IEEEbiography}[{\includegraphics[width=1in,height=1.25in,clip]{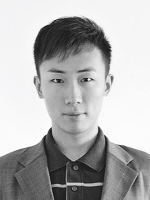}}]{Liang Du}
 received the B.S. degree from Harbin Engineering University, China, in 2016. He received the M.S. degree from the University of Chinese Academy of Sciences, China, in 2019. He is currently pursuing the Ph.D. degree at Fudan University, China. His current research interests include 3D computer vision, transfer learning, and autonomous driving. He has contributed several papers in top conferences, such as CVPR, ICCV, and ECCV.
\end{IEEEbiography}

\vspace{-16 mm} 

\begin{IEEEbiography}[{\includegraphics[width=1in,height=1.25in,clip]{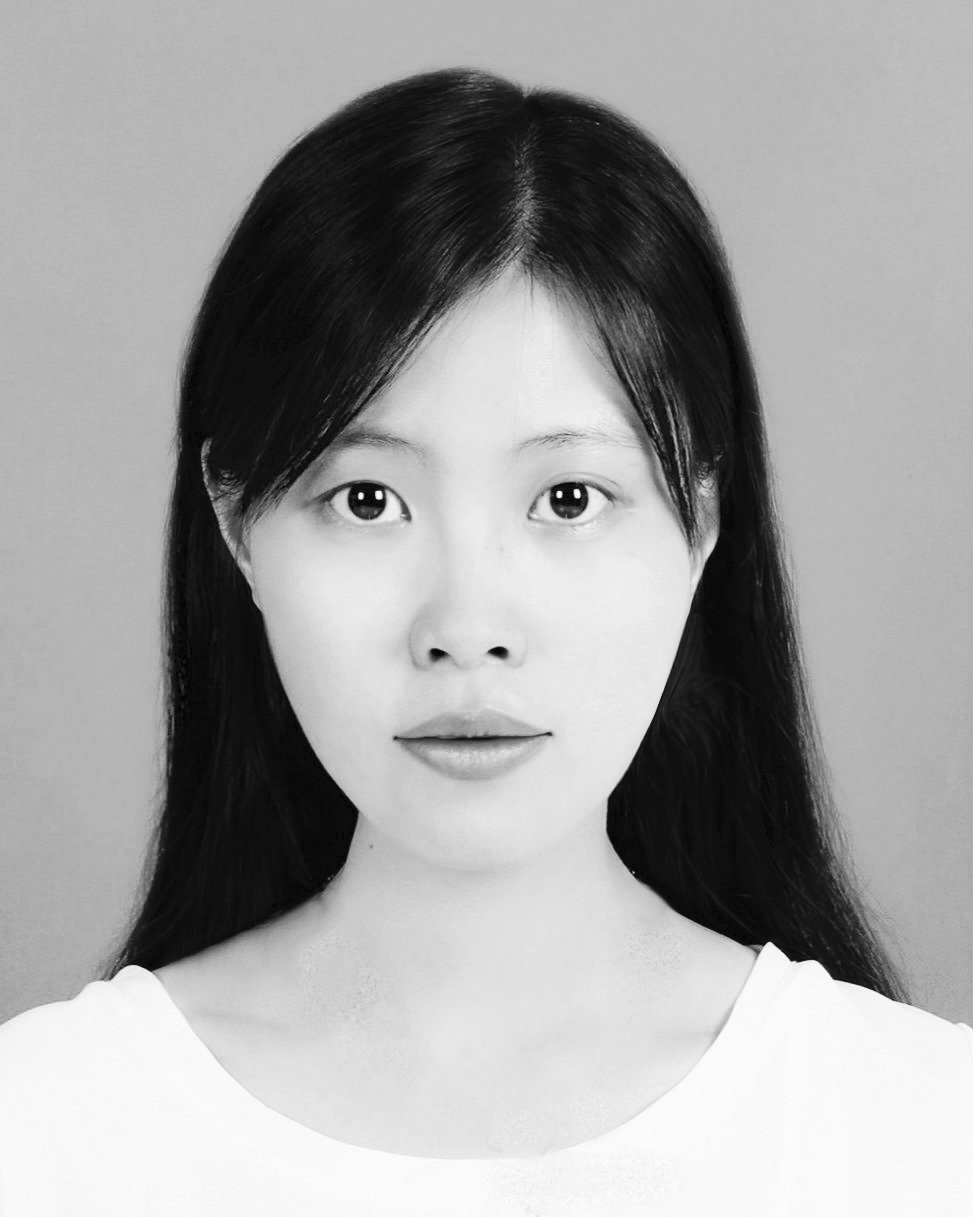}}]{Xiaoqing Ye}
 received the Ph.D. degree in information and communication engineering from the University of Chinese Academy of Sciences, China, in 2019. She is currently with the department of computer vision technology, Baidu as a Senior Engineer. Her current research interests include 3D computer vision and autonomous driving. She has contributed several papers in top conferences, such as CVPR, ECCV, ICCV and AAAI.
\end{IEEEbiography}

\vspace{-16 mm} 

\begin{IEEEbiography}[{\includegraphics[width=1in,height=1.25in,clip]{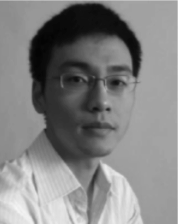}}]{Xiao Tan}
 received the Ph.D. degree in computer vision from the University of New South Wales, Sydney, in 2014. His research interests include computer vision, pattern recognition, and image processing. He is currently with the department of computer vision technology, Baidu as a Senior Engineer, leading the tech-team to develop visual systems for AI-city and autonomous driving. He has contributed 10+ papers in top conferences and journals, such as CVPR, ECCV, ICCV, and TIP.
\end{IEEEbiography}

\vspace{-16 mm} 

\begin{IEEEbiography}[{\includegraphics[width=1in,height=1.25in,clip]{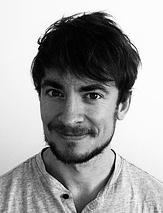}}]{Edward Johns}
 received the B.A. and M.Eng. degrees from the University of Cambridge in 2006 and 2007, and the PhD degree from Imperial College London in 2014. He is currently Director of the Robot Learning Lab at Imperial College London, where his research focusses on the intersection between robotics, computer vision, and machine learning. He has contributed 30+ papers in top conferences and journals, such as NeurIPS, CVPR, ICCV, ECCV, and IJCV.
\end{IEEEbiography}

\vspace{-16 mm} 

\begin{IEEEbiography}[{\includegraphics[width=1in,height=1.25in,clip]{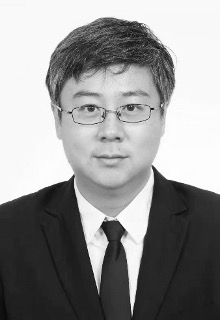}}]{Bo Chen}
 received the B.S., and M.S. degrees from Jilin University, China. He is the general manager of FAW (Nanjing) Technology Development Co., Ltd, and experts in artificial intelligence. He is currently in charge of the development of autonomous driving based on AI algorithm.
\end{IEEEbiography}

\vspace{-16 mm} 

\begin{IEEEbiography}[{\includegraphics[width=1in,height=1.25in,clip]{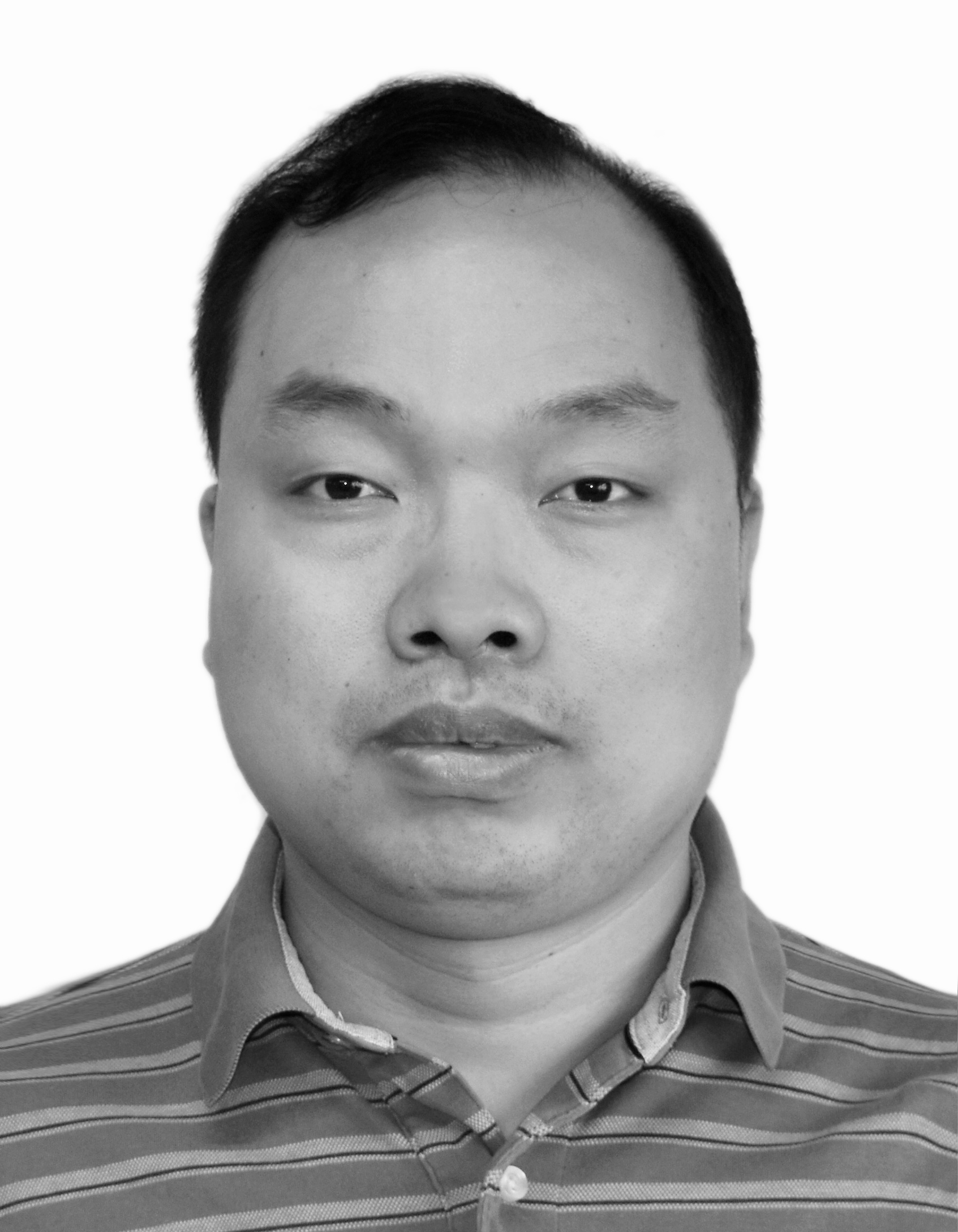}}]{Errui Ding}
 received Ph.D. degree from Xidian University in 2008 and currently is the director of Computer Vision Technology Department (VIS) of Baidu Inc. In recent years, he has published tens of papers on top-tier conferences and was awarded Best Paper Runner-up at ICDAR 2019. He co-organized several competitions and workshops at recent ICDAR and CVPR. He is also a member of CSIG, CCF and CAAI.
\end{IEEEbiography}

\vspace{-16 mm} 

\begin{IEEEbiography}[{\includegraphics[width=1in,height=1.25in,clip]{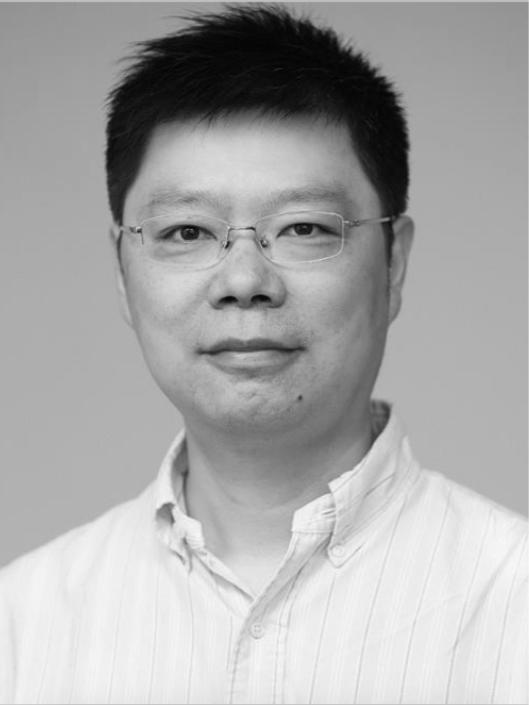}}]{Xiangyang Xue}
 received the B.S., M.S., and Ph.D. degrees in communication engineering from Xidian University, Xian, China, in 1989, 1992, and 1995, respectively. He is currently a professor of computer science with Fudan University, Shanghai, China. His research interests include computer vision, multimedia information processing, and machine learning.
\end{IEEEbiography}

\vspace{-16 mm} 

\begin{IEEEbiography}[{\includegraphics[width=1in,height=1.25in,clip]{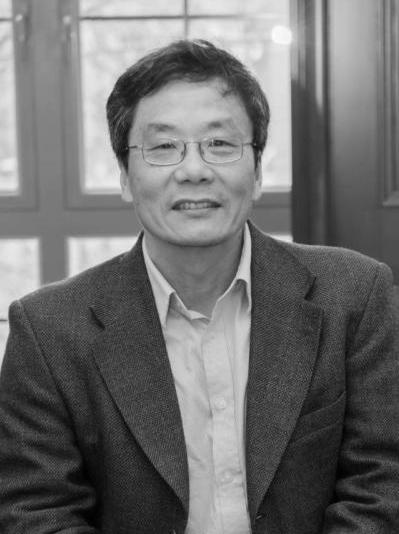}}]{Jianfeng Feng}
 received the B.S., M.S., and Ph.D. degrees from the Department of Probability and Statistics, Peking University, China. He is the chair professor of Shanghai National Centre for Mathematic Sciences and the Dean of Brain-inspired AI Institute in Fudan University. He has been developing new mathematical, statistical, and computational theories and methods to meet the challenges raised in neuroscience and mental health researches.
\end{IEEEbiography}

\end{document}